\documentclass[
  english,        % define the document language (english, german)
  onecolumn ,      % use onecolumn or two-column layout
]{tumarticle}
\usepackage[utf8]{inputenc}

\usepackage[utf8]{inputenc}
\usepackage{amsmath}
\usepackage{tabularx}
\usepackage{makecell}
\usepackage{booktabs}  % table rulers 
\usepackage[T1]{fontenc}
\usepackage{lmodern}
\usepackage{cite}
\usepackage{listings}
\usepackage{xcolor}
\definecolor{codegreen}{rgb}{0,0.6,0}
\definecolor{codegray}{rgb}{0.5,0.5,0.5}
\definecolor{codepurple}{rgb}{0.58,0,0.82}
\definecolor{codered}{RGB}{255, 67, 54}
\definecolor{backcolour}{rgb}{0.95,0.95,0.95}
 
\usepackage{amsfonts}
\usepackage{amsmath}
\usepackage{amsthm}  % theorems, definition etc.
\usepackage{amssymb}
\usepackage{aligned-overset}

\usepackage[ruled,linesnumbered,inoutnumbered]{algorithm2e}
\SetKwFunction{Range}{range}
\SetKw{KwTo}{in}\SetKwFor{For}{for}{\string:}{}

\SetCommentSty{mycommfont}

\usepackage{fontawesome5}

\usepackage{tcolorbox}

\usepackage{graphicx}
\graphicspath{{figures/}}
\usepackage{caption}
\usepackage{subcaption}
\newtheorem*{probdefinition*}{Problem Definition}

\usepackage{lipsum}

\newcommand{\N}{\ensuremath{\mathbb{N}}}

\renewcommand{\L}{\mathcal{L}}

\newcommand{\R}{\ensuremath{\mathbb{R}}}
\newcommand{\C}{\ensuremath{\mathbb{C}}}

\newcommand{\Zc}{\mathcal{Z}}

\newcommand{\X}{\ensuremath{\mathcal{X}}}
\newcommand{\w}{ \boldsymbol{\omega}}

\newcommand{\x}{\bm{x}}

\newcommand{\E}{\mathbb{E}}

\newcommand{\im}{\mathrm{i}}

\newcommand{\Y}{ \mathcal{Y}}

\newcommand{\M}{\mathcal{M}}
\newcommand{\Mh}{\widehat{\mathcal{M}}}

\newcommand{\K}{\mathbf{K}}

\newcommand{\HH}{\mathcal{H}}

\newcommand{\xb}{\mathbf{x}}
\newcommand{\cb}{\mathbf{c}}
\newcommand{\hb}{\mathbf{h}}
\newcommand{\yb}{\mathbf{y}}
\newcommand{\zb}{\mathbf{z}}

\newcommand{\Ab}{\mathbf{A}}
\newcommand{\xcb}{\mathbf{X}}
\newcommand{\ycb}{\mathbf{Y}}

\title{Non-intrusive surrogate modelling using sparse random features with applications in crashworthiness analysis}

\author[affil=1 2, email=maternus.herold@tum.de]{Maternus Herold}
\author[affil=1 3 4]{Anna Veselovska}
\author[affil=2]{Jonas Jehle}
\author[affil=1 3 4]{Felix Krahmer}

\affil[mark=1]{\theDepartmentName, \theUniversityName}
\affil[mark=2]{BMW Group}
\affil[mark=3]{Munich Data Science Institute}
\affil[mark=4]{Munich Center for Machine Learning}
\date{\today}
\begin{document}

\maketitle

\begin{abstract}
  Efficient surrogate modelling is a key requirement for uncertainty quantification in data-driven scenarios. In this work, a novel approach of using Sparse Random Features for surrogate modelling in combination with self-supervised dimensionality reduction is described. The method is compared to other methods on synthetic and real data obtained from crashworthiness analyses. The results show a superiority of the here described approach over state of the art surrogate modelling techniques, Polynomial Chaos Expansions and Neural Networks.
\end{abstract}

\sloppy
% -----------------------------------------------------------------------------
\section{Introduction}\label{sec:introduction}

Computer simulations are at the core of today's engineering craft as they allow for the design and analysis of new products in silico, rather than building those artifacts with hardware. The automotive industry, to name one, is heavily relying on computer simulations to test the crashworthiness of its vehicles during development \cite{jonasjehlephd, Jehle2021, jehle2021enabling, crashworthinesseichmueller}. Further, having an accurate computational model is valuable to gain a comprehensive understanding of how the object under observation behaves, e.g., a car's body during impact. While such information is easily obtainable from the simulation, the results have to be treated with caution due to uncertainty in the parameters and the model design \cite{sullivan2013, uqsoize2017uncertainty, uqghanem2017handbook, uqsmith2013uncertainty,uqabdar2021review}. As the computer simulations tend to have a long runtime, combined with large input dimensions \cite{jonasjehlephd, Luethen2020}, quantifying the uncertainty numerically is challenging. 
A best practice approach to overcome this difficulty is to learn a data-adaptive surrogate model approximating the simulations that is  computationally cheaper, and hence allows for a larger number of evaluations with different parameter settings. 

%surrogate models are required to approximate the simulation, allowing many evaluations for different parameter settings.

In this work, the relations between inputs and outputs are considered as black-box evaluations of the simulation, i.e. the functions of the generating simulation cannot be altered. A common strategy in such non-intrusive settings is to use readily available data from previously run simulations to learn the surrogate models, framing a data-driven approach \cite{lataniotis2020extending,sullivan2013}. Due to the long run times, the available datasets are small, compared to the high dimensions, leading to the curse of dimensionality when following a data-driven approach. 

% todo: finde more references 
State-of-the-art methods use the sparsity-of-effects principle to reduce the model's basis \cite{Luethen2020,sullivan2013}, apply dimensionality reduction in a self-supervised fashion \cite{lataniotis2020extending} or use small neural networks, which implicitly incorporating the two aforementioned approaches \cite{nnsmtripathy2018deep,nnsmpfrommer2018optimisation}. In this paper, we propose an alternative approach based on 
adapting  sparse random feature expansions (SRFE) \cite{hashemi2021generalization} to surrogate modelling. We compare this algorithm with state-of-the-art methods on computational benchmark data and real crashworthiness data. 
To the best of our knowledge, our study is the first successful demonstration of using Sparse Random Features in the field of data-driven surrogate modelling for uncertainty quantification (UQ).

\subsection{Data-Driven Surrogate Modelling for Uncertainty Quantification}\label{subsec:sm-for-uncertainty-quantification}

In the context of uncertainty quantification (UQ), a physical or computational  model of a system can be seen as a black box that performs the mapping 
\begin{equation*}
    \ycb= \M(\xcb)
\end{equation*}
where $\xcb$ is a random vector that parameterizes the uncertainty of the input through a joint probability density function (PDF) $\rho_{\xcb}$ and $\ycb$ is the corresponding random vector of model response. The leading goal of  UQ is to propagate the uncertainties from $\xcb$ to $\ycb$ thought $\M$, which may require several thousands of evaluations of the physical model $\M$ for different realizations of $\xb$ of $\xcb$. However, most models used in applications have high computational costs per model run, and consequently cannot be used directly \cite{lataniotis2020extending}. To alleviate the computational burden, surrogate modelling has become a driving tool in the area of $UQ$.

A surrogate model $\Mh_{\hb,c}$ is a computationally inexpensive approximation of the true model $\Mh$ such that 
\begin{equation}\label{eq:base-model-approx}
    \mathcal{M}(\xcb) \approx
    \Mh_{\hb, c}(\xcb),
\end{equation}
where the hyper-parameters $\hb$ encode fixed design  features and the parameters $c$ specify the configuration of the surrogate model; they are usually calibrated based on the evaluations of the full model on a training set $\X_{\text{exp}}:=\{\xb^{\text{exp}}_1, \ldots, \xb^{\text{exp}}_n\}$.
% , and change all the following notation for $\X_{\text{exp}}$ and $\X_{test}$}
The set $\X_{\text{exp}}$ is called \textit{experimental design} and the set of the corresponding evaluations $\Y_{\text{exp}}:=\left\{\M(\xb^{\text{exp}}_1), \ldots, \M(\xb^{\text{exp}}_n) \right\} $ is referred to as \textit{model response}.

In the literature, two types of surrogate  models have received particular attention. 
Firstly,  surrogate models based on polynomial chaos expansions take the form $\Mh_{\mathcal{A},c}(\xcb)=\sum_{\alpha\in \mathcal{A}} c_{\alpha}\xi_\alpha(\xcb)$, where $\xi_\alpha(\xcb)$  are multivariate polynomials orthonormal w.r.t. some distribution $\rho_{\xcb}$ and the hyper-parameters $\hb=\mathcal{A}\subset \N^d$ characterize the model by specifying the active polynomial elements used in the expansion \cite{lataniotis2020extending}. Secondly,  surrogate models based on neural networks, 
%are given by a neural network configuration of a certain architecture.  Here 
$\hb$ encodes the number of layers and their widths, while $c$ captures the data-adapted weights. In this paper, we will focus on a third type of surrogate models, the so-called \textit{sparse random feature expansion},  which will be described in the following section.

%To choose among the pool of surrogate models and hyper-parameters defining them,
The quality of  a surrogate model can be measured by the relative generalization error $E$ defined as 
\begin{equation}\label{eq:cont:error}
    E=\E \left[ (\ycb - \Mh_{\hb}(\xcb))^2\right]/Var[\ycb].
\end{equation}
%which can be used in the calibration of parameters $\theta$. 
To estimate this error for an unknown distribution, one considers the empirical relative generalization error on a  \textit{validation set } $\X_{\text{test}}= \{\xb^{\text{test}}_1, \ldots, \xb^{\text{test}}_{N_{\text{test}}}\}$ with model response $\Y_{\text{val}}= \{\M(\xb_1), \ldots, \M(\xb_{N_{\text{test}}})\}$, which, in line with machine learning terminology, is often referred to as \textit{generalization error} and given  by
\begin{equation}\label{eq:emp:error}
\widehat{\epsilon}(\X_{\text{test}},  \Y_{\text{test}}, \Mh)=\frac{\sum_{i=1}^{N_{\text{test}}} (y^{\text{test}}_i- \Mh(\xb^{\text{test}}_i))^2}{\sum_{i=1}^{N_{\text{test}}}(y^{\text{test}}_i- \mu_y)^2}.
\end{equation}
Here $ y^{\text{test}}_i: = \M(\xb^{\text{test}}_i)$ and $\mu_y=\frac{1}{N_{\text{test}}} \sum_{i=1}^{N_{\text{test}}} y^{\text{test}}_i$ is the sample mean of the validation set for the model response. 
The corresponding quantity evaluated on the experimental design $\X_{\text{exp}}$, i.e., $\widehat{\epsilon}(\X_{\text{exp}},  \Y_{\text{exp}}, \Mh)$, is often referred to as training error.

\subsection{Outline}

This paper is structured in the following manner. In  Section~\ref{sec:literatur-and-ingredients}, we  describe computational tools for data-driven surrogate modelling for UQ. 
% we will describe the general setting of data-driven surrogate modelling for UQ in \Cref{subsec:sm-for-uncertainty-quantification}. 
Then, we present the a novel method of self-supervised surrogate modelling with low order sparse random feature expansion (LOS-RFE) in Section~\ref{sec:self-supervised-surrogate-modeling}, including a discussion of the used optimization components and implementation.  %and the adaptions made to SRFEs. 
In Section~\ref{sec:numerical-experiments}, the presented LOS-RFE method is experimentally compared to a combination of dimensionality reduction and polynomial chaos expansion \cite{lataniotis2020extending} and neural networks, which are both state-of-the-art surrogate modelling techniques. The numerical comparison is done both on synthetic data and on real data obtained from studies on crashworthiness. 

\section{Background on Computational Tools}\label{sec:literatur-and-ingredients}

In this paper, we propose a new method for surrogate modelling in high dimensions that emerges through  a combination of sparse random feature expansion and dimensionality reduction to self-supervised learning. In this section, we provide some background on each of these topics. After discussing  sparse random feature expansions in  Section~\ref{sec:srfe}, we review methods for  dimensionality reduction in Section~\ref{sec:dimred} and describe our optimization method of choice in Section~\ref{sec:pso}.
%The proposed approach searches over a set of hyperparameters, such as lower dimensions, optimizing the dimensionality reduction's parameters by minimizing the surrogate's loss on the training data. 

%Here, we describe the recently introduced sparse random feature expansion \cite{hashemi2021generalization} as a surrogate modeling method and propose several modifications that allow reaching better performance in the self-supervised regime. 

\subsection{Sparse Random Feature Expansion via the LASSO}\label{sec:srfe}

% Given an experiment design $\X=\{\xb_1, \ldots, \xb_n\}$ drawn from a certain distribution and the corresponding  model responses $\Y=\left\{\M(\xb_1), \ldots, \M(\xb_n) \right\} $, we aim at approximating the computational model

% $\M$ using 
The idea of the
\textit{sparse random feature expansion}  (SRFE) \cite{hashemi2021generalization} is to  fit the target model $\M$ by sparse linear combinations of elements  of a sufficiently rich randomized function family.  
To be more precise, one fixes a bivariate function $\phi$  and considers the random collection of functions $\phi(\cdot; \w_i)$, $i=1, \ldots, R$, with weight vectors $\w_i \in \R^d$ drawn independently at random from some probability distribution $\rho(\w)$. These functions $\phi(\cdot; \w_1), \ldots, \phi(\cdot; \w_R)$ are called \textit{random features} \cite{randomfeatforlargescalekernelmachines}.
%A random feature expansion,  the model $\M$ needs to be approximated by a linear combination of the random features  $\phi(\cdot; \w_1) \ldots \phi(\cdot; \w_R)$ given the data $\Y$. Mathematically, it 
Random features give rise  to surrogate models $f^{\star}$ of the form
\begin{equation}\label{eq:sm-rf}
  f^{\star}(\xcb):= \sum_{j=1}^R c_j \, \phi(\xcb; \w_j).
    \end{equation}
%with random intup vector $\xcb$ and a coefficient vector $\cb:=(c_1, \ldots, c_N) \in \R^R$,  
Consequently, given a model $\M$, the aim is to find coefficients $\cb:=(c_1, \ldots, c_N) \in \R^R$ such that the resulting \textit{ random feature expansion} $f^*$ approximates $\M$ \cite{randomfeatforlargescalekernelmachines}, that is,
\begin{equation}\label{eq:fun-ex-rf}
 \M(\xcb)=f^{\star}(\xcb)+\varepsilon,
\end{equation}
with some sufficiently small error term $\varepsilon$. 
Popular choices for $\phi(\cdot; \w)$ include 
\begin{itemize}
    \item random Fourier features: $\phi(\cdot; \w)=\exp (\im\langle \xb, \w\rangle)$
    \item  random trigonometric features: $\phi(\cdot; \w)=\cos (\langle \xb, \w\rangle)$ and $\phi(\cdot; \w)=\sin (\langle \xb, \w\rangle)$
    \item  random ReLU features:  $\phi(\cdot; \w)=\max\{ \langle \xb, \w\rangle, 0\}$.
\end{itemize}
%Representation  \eqref{eq:fun-ex-rf} is commonly referred to as \textit{ random feature expansion} (RFE) of $\M$ . 

It was noted  in \cite{hashemi2021generalization} that the generalization ability of the surrogate model may improve when enforcing in addition that the coefficient vector $c$ is sparse, that is, it has only a small number of non-vanishing entries. This is also the approach that we will be following in this paper. More precisely, we consider the so-called \textit{ low order sparse RFE} (LOS-RFE) \cite{hashemi2021generalization} that promotes  \textit{ bi-level sparsity} in the expansion \eqref{eq:fun-ex-rf}.
The first sparsity level is motivated by the fact that in many applications  $\M$ is well-approximated by a small subset of  the random features $\phi(\cdot; \w_1) \ldots \phi(\cdot; \w_R)$. Therefore, one aims for $\cb$ with only few non-zero entries,  which can be done by minimizing $\|\cb\|_1$ with respect to the given model response $\Y$. 
%the residual norm $\|A\cb - \yb\|_2$ with  
%in terms of dictionary matrix $\Ab \in \C ^{m\times N} $ with entries $a_{jk}=\phi(x_k, \omega_j)$ as $A\cb \approx \yb$ allows to use sparse regression methods for determining $\cb$. In this case, one looks for $\cb$ that minimizes the residual norm $\|A\cb - \yb\|_2$ under the constraint that only few entries of $\cb$ are non-zero.
 
Secondly, we enforce the \textit{ sparsity-of-effects} or Pareto principle, which states that most real-world systems are dominated by a small number of low-complexity interactions. Consequently, we also promote sparsity for the feature weights $\w$.   Mathematically, one can model this principle by the so-called \textit{ order-$q$ functions}. We call a function $f: \R^d \to \R$  of order-$q$ of at most $K$ terms $q< d$, if there exist $K$ functions $g_1, \ldots, g_K: \R^q\to \R$ such that 
\begin{equation}\label{eq:order-q-fun}
    g(x_1, \ldots, x_d)= \frac{1}{K} \sum_{j=1}^K g_j(x_{j_1}, \ldots, x_{j_q}). 
\end{equation}
Since for the target model $\M$, the set of active variables is not known in advance, to incorporate the potential coordinate sparsity into the weights $\w_i$,  one randomly draws $R=n { d \choose q}$ $q$-sparse features $\w_j$ with nonzero entries distributed w.r.t. $\rho_q(\w)$ for each subset of size $q$ from the index set $\{1,\ldots, d\}$, for more details see \cite{hashemi2021generalization}.

%%%%%%%%% moved from the next Section %%%%%%%%%%%%%%%%%

\begin{algorithm}
    \caption{Sparse Random Feature Expansion via the LASSO} \label{alg:srfe}
    \SetKwInOut{input}{input}
    \SetKwInOut{output}{output}    

    \input{observed data $\{(\xb_i, y_i)\}^{m}_{i=1}$, parametric basis function $\phi_{\w}(\mathbf{x}) = \phi(\langle \mathbf{x}, \w \rangle)$, feature sparsity $q$, probably measure $\rho$ and a stability parameter $\eta$}

    % return optimal weights and min. error 
    \output{surrogate model:  $f^{\star}(\mathbf{x}) = \sum^R_{j=1} c^{\ast}_j \phi_{\w_j}(\mathbf{x})$}

    \BlankLine  % empty line 

    % init variables 
    \tcc{draw $q$-sparse feature weights}
    $\Omega = \{ \w_j \sim \rho_q(\w) \ \vert \ \forall j \in [1, \dots, R] \}$
    
    \BlankLine  % empty line

    \tcc{construct random feature matrix}
    $ \mathbf{A} \in \mathbb{R}^{m \times R} $ s.t. $ a_{kj} = \phi_{\w_j}(\mathbf{x}_k) $

    \BlankLine  % empty line

    \tcc{solve the Lasso problem to obtain the coefficients}
    $ \mathbf{c}^{\ast} = \underset{\mathbf{c}}{\operatorname{arg min}} \; \lambda \vert\vert \mathbf{c} \vert\vert_1 + \vert\vert \mathbf{Ac - y} \vert\vert_2^2$

\end{algorithm}

 To track the coefficient vector $\cb$ numerically, one can proceed as follows. Let  $\Ab \in \C ^{m\times R} $ be given by $a_{jk}=\phi(\xb_k, \w_j)$ and consider the vector $y=(y_1,\ldots, y_m)$ with $y_i=\M(\xb_i)$ representing the vectorized data $\Y$. Then approximating $\M$ by $f^\star$ as in \eqref{eq:fun-ex-rf} with sparse $\cb$ is equivalent to finding $\cb$ that minimizes the residual norm $\|A\cb - \yb\|_2$ under the constraint that only a few entries of $\cb$ are non-zero. % There is a pool of sparse regression solvers that can used in the purpose of finding such $\cb$, see \cite{luthen2020sparse}. 
In \cite{hashemi2021generalization}, it has been proposed to track $\cb$ via the BP-denoising problem. We observed, however, that in the context of the problems studied in this paper, solving a BP-denoising problem takes too long to be used in a self-supervised loop where optimization is run several hundreds of times. 
%Next to the parameters for dimensionality reduction, also the targeted latent dimension $k$ as well as model specific hyperparameters such as the number of random features $R$ and $q$ are searched, adding an additional layer to the optimization problem. . 
% Also, in the self-supervised process, %as described in Algorithm~\ref{alg:srfe},
% we want the hyperparameters of the surrogate model $f^\star$, such as the number of features $R$, sparsity $q$ to be tuned
Hence, a more efficient optimization is required, and to  find the coefficient vector $\cb$ we propose to use the following LASSO formulation 
\begin{align}\label{eq:lasso-for-later-phase}
    &\underset{\mathbf{c} \in \mathbb{R}^d}{\operatorname{arg min}} \ \lambda \| \mathbf{c} \|_1 + \| \mathbf{Ac} - \mathbf{y} \|^2_2. 
\end{align}   
 
 In this formulation, the hyperparameter $\lambda$ balances the sparsity inducing regularizer and the data fidelity. Indeed, if $\lambda$ is well chosen, \eqref{eq:lasso-for-later-phase} will reduce coefficients to zero, yielding a sparse solution.

From the theoretical perspective, it has been shown in \cite{randomfeatforlargescalekernelmachines}  that if the model $\M$ belongs to a reproducing kernel Hilbert space (associated with the choice of the feature functions), then the model can be uniformly approximated by the RFE as in \eqref{eq:sm-rf} with any accuracy $\varepsilon$ in the case when the number of random features $R$ is chosen of order $1/\varepsilon^2$. For the LOS-RFE, it has been shown that low-order functions can be approximated with the accuracy $\varepsilon$ with a significantly lower number of random features $R$, see \cite{hashemi2021generalization} for more details.   We believe that this improvement is key for obtaining an advantage in comparison to the neural-network-based methods and Polynomial Chaos Expansion (PCE).

\subsection{Dimensionality Reduction via Kernel Principal Component Analysis}\label{sec:dimred}

A common challenge in learning a surrogate model is that the experimental design 
${\X=\{\xb_1, \ldots, \xb_N\}}$ is rather high-dimensional, so a good surrogate model will need to have a large number of parameters, which leads to computational bottlenecks. At the same time,  the experimental design  is often intrinsically low-dimensional, that is, all data points lie close to a lower-dimensional manifold.  A common strategy is to first identify this lower-dimensional structure and then learn the surrogate model only on this set of interest.

Mathematically, this dimensionality reduction (DR) corresponds to a non-linear mapping $\tau:\R^d\to \R^k$ with  $k \ll d$. We denote the image of $\X$ under this map  by $ \Zc =\{\zb_i= \tau(\xb_i),\, i=1,\ldots, N\}\subset \R^k$. The aim is to choose the parameter $k$ close to the dimension of the underlying manifold; consequently $k$ is often referred to as the “intrinsic dimension” of $\X$.

In this work,  we use \textit{ kernel principal component analysis} (KPCA) \cite{kpcaSchoelkopf, mika1998kernel, Hastie2009, Bishop2006}, which is a non-linear variant of principal component analysis (PCA), where the data is first mapped to a higher dimensional or potentially even infinite-dimensional Hilbert space $\HH$ via a feature map ${\Phi:\R^d \to \HH }$, and the PCA is applied. 

 Analogously to the classical PCA case, to directly compute the principal components  of $\Phi(\X)$ one  would need to perform the eigen-decomposition of the sample
covariance matrix ${C_{\Phi(\X)}=\frac{1}{N}\sum_{i=1}^N\Phi(\xb_i) \Phi(\xb_i)^*}$. 
Such a direct computation, however, is, in general, not feasible and the dimension of $\HH$ is typically very large or even infinite.  
This problem is by-passed by observing that each eigenvector $\mathbf{ v}_k$ belongs to the span of the samples $\Phi(\xb_1),\ldots, \Phi(\xb_M)$,  therefore scalar coefficients $\alpha_k^i$ exist \cite{kpcaSchoelkopf}, such that each eigenvector $\mathbf{v}_m$ can be expressed as the following linear combination
\begin{equation}\label{eq:eig-vectors-kpca}
    \mathbf{ v}_m= \sum_{i=1}^N\alpha^m_i \Phi(\xb_i), \quad m=1, \ldots, N. 
\end{equation}
Consequently, 
\begin{equation}\label{eq:eig-vectors-kpca-and-matrix}
   C_{\Phi(\X)} \mathbf{ v}_m=  \sum_{j=1}^N  \sum_{i=1}^N \alpha^m_i \Phi(\xb_j) \Big(\Phi(\xb_j) \cdot \Phi(\xb_i)\Big). 
\end{equation}
Thus, if one considers the kernel function $\kappa: \R^d\times \R^d \to \R$  given by 
 \begin{equation}\label{eq:kern:gen}
\kappa(\mathbf{x,y})= \Phi(\xb)\cdot \Phi(\yb),
 \end{equation}
one can apply the so-called kernel trick, which refers to the observation
that, if the access to $\HH$ only takes place through inner products, then there is no need to explicitly compute $\Phi$. Rather, the result of the inner product can be directly calculated using $\kappa(\cdot,\cdot)$.
With this observation, one can reformulate the eigenvalue problem for the matrix  $C_{\Phi(\X)}$ as the following eigenvalue problem
\begin{equation}\label{eq:eig-vec-kernel-kpca}
\K\alpha^m=\lambda_i\alpha^m, \quad m=1, \ldots, N
\end{equation}
for the kernel matrix $\K$  given by $  \K_{ij}=\kappa(\xb_i, \xb_j)$ and the coefficient vector $\alpha^m$ in representation \eqref{eq:eig-vectors-kpca}. The $\alpha^m$ can then  be converted back to the eigenvectors $\mathbf{v}_{m}$. 

This gives rise to a dimensionality reduction method (see, e.g., \cite{kpcaSchoelkopf}) where each data point  projected to the $k$ leading principal axes   ${\mathbf{V}_k:=\{
\mathbf{v}_m, \; m = 1, \ldots,k\}}$.
That is, each $\xb_i$ is represented by a vector $\zb_i = (\zb_i^1, \dots, \zb_i^k)^T$ with entries
\begin{equation}
    \zb^m_i= \Phi(\xb_i)^T \mathbf{v}_m=\sum_{j=1}^N\alpha_i^mk(\xb_i,\xb_j). 
\end{equation}
Note that this method does not explicitly use the feature map and can hence be defined also for more general classes of kernels.

In this paper, we  use KPCA with a nonisotropic Gaussian kernel parameterized by a vector $\theta \in \R^d$ as given by 
\begin{equation}\label{eq:gaus-kernel} %\theta^{2}_{\text{step},p}}
    % \label{eq:gaussian-kernel}
    \kappa_{\theta}(\mathbf{x,y}) = \exp \left( - \frac{1}{2}
\sum_{i=1}^d\frac{1}{\theta_i^2}\big(\xb_i-\yb_i\big)^2 \right),
\end{equation}
 and denote the corresponding dimensionality reduction map by $\tau^k_{\theta}$ 
 %the projected data  $\Zc$ can be expressed as ${\Zc= \tau^k_{\theta}(\X)}$, 
 with $\theta$ and $k$ indicating the kernel parameters and  the reduced dimension, respectively.

\subsection{Particle Swarm Optimization}\label{sec:pso}

The aim of our approach is choose the parameter $\theta$ in the Gaussian kernel in a data-adaptive way. To optimize the choice of $\theta$, we will use particle swarm optimization (PSO)\cite{Poli2007}, a gradient-free metaheuristic to minimize a given loss-function $\L: \mathcal{D}\subset \R^m\to\R$. 

The idea behind the PSO metaheuristic, as the name suggests, is to use a large number of points in the domain $\mathcal{D}$ called particles, a particle swarm, in order to explore the landscape of the loss function in an iterative way. As a starting point, the particles are randomly initialized, and then each particle explores the co-domain of the function driven by the difference between the values of $\L$ at its own position and at the positions of the other particles. 
After a certain number iterations, the method outputs the minimum point ever explored by this swarm of particles. 

More rigorously, assume that we have $P$ particles and the position of a particle $p \in \{1,\dots, P\}$ at iteration $t$ is denoted by $\widehat{\theta}_p(t) \in \R^m$ and its velocity is denoted by $V_p(t)\in \R^m$.
This velocity is recomputed in each iteration step as a combination of three contributions
\begin{itemize}
\item The velocity $V_p(t)$ of the previous iteration, weighted by a learning rate $r\in [0,1]$.
\item A cognitive term driving the particle towards the location of minimal loss  $\widehat{\theta}_{i, p}^{\, pbest}$  it has previously encountered, weighted by a constant $c_{cognitive}$ and a random positive rescaling $u_1(t)\sim \mathcal{U}[0,\xi_{1}]^d$ for some $\xi_1\in \R$.
\item A social term driving the particle towards the location of minimal loss $\widehat{\theta}_{i}^{\, gbest}$  that has been previously encountered by any particle in the swarm, weighted by a constant $c_{social}$ and a random positive rescaling $u_2(t)\sim \mathcal{U}[0,\xi_{2}]^d$ for some $\xi_2\in \R$.
\end{itemize}
In formulas, this update reads as 
\begin{equation}
    V_p(t+1)=  r \cdot V_{p}(t) + c_{\text{cognitive}} \cdot u_{1}(t) \left[ \widehat{\theta}_{i, p}^{\, pbest} - \widehat{\theta}_{p}(t) \right] + c_{\text{social}} \cdot u_{2}(t)\left[  \widehat{\theta}_{i}^{\, gbest} - \widehat{\theta}_{p}(t) \right].
\end{equation}
The random weights $u_1$, $u_2$ are often called acceleration coefficients \cite{Poli2007} and the underlying parameters  $\xi_{1,2}$ determining the magnitude of the random force is usually set equal to $2$ \cite{Poli2007}. 
With this velocity vector, the position  $\widehat{\theta}_p(t)$ of particle $p$ is updated as 
\begin{equation}\label{eq:pso-update}
    \widehat{\theta}_p(t+1)= \widehat{\theta}_p(t)+ V_p(t+1).
\end{equation} 
We will denote the output of the PSO algorithm with starting points  $\{\widehat{\theta}_i \}_{i=1}^P$ and  loss function $\L$ by ${\textbf{ PSO}}( \L,  \{\widehat{\theta_i}\}_{i=1}^P )$. %and outputs the minimizer $\widehat{\theta}^\star$ of $\L$. 

\begin{figure}
    \centering
    \subfloat[\label{fig:ackley-state-1}]{
        \includegraphics[width=0.3\linewidth]{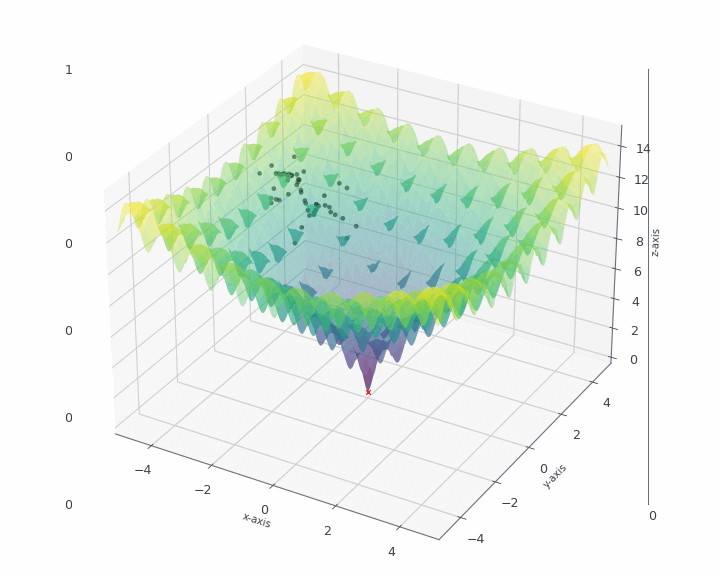}
    }
    \subfloat[\label{fig:ackley-state-2}]{
       \includegraphics[width=0.3\linewidth]{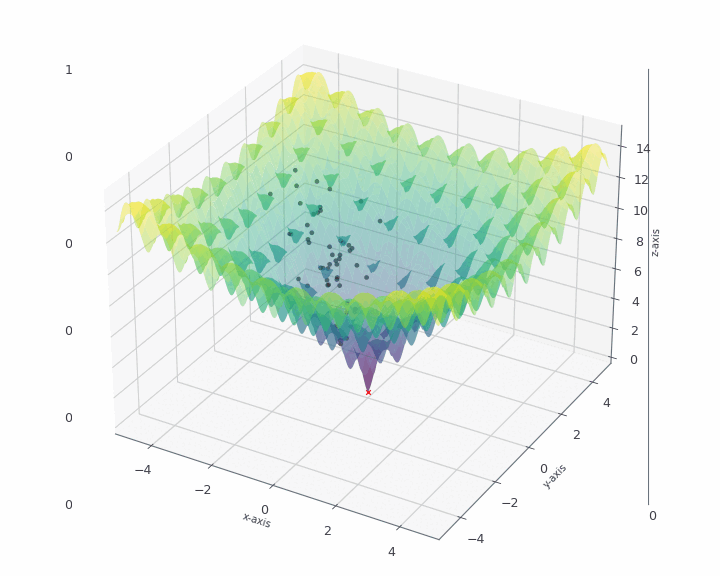}
    }
    \subfloat[\label{fig:ackley-state-3}]{
        \includegraphics[width=0.3\linewidth]{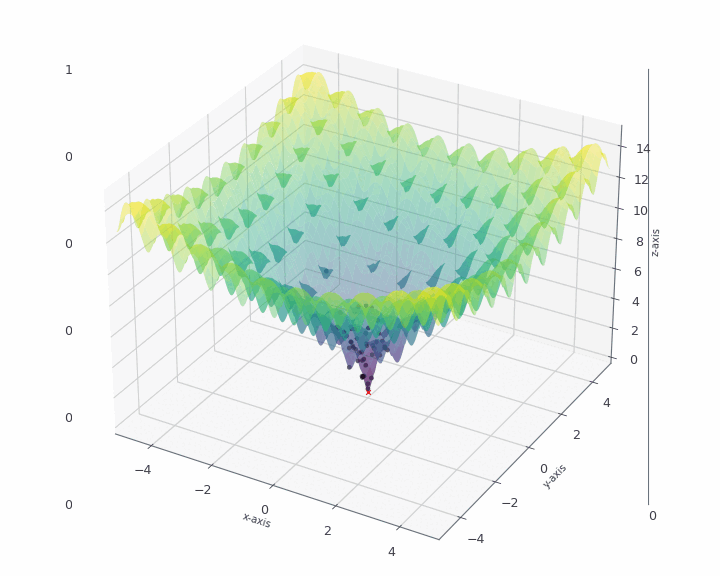}
    }
    
    \caption[Particle Swarm Optimization on Ackley benchmark function]{ The PSO algorithm is applied to the popular Ackley function. The different particles used are marked as black dots.  Fig. \ref{fig:ackley-state-1} shows locations of the particles shortly after the start of the optimization. Fig. \ref{fig:ackley-state-2} indicates the positions of particles after a couple of iterations, and Fig.\ref{fig:ackley-state-3} shows the final iteration stage.   As one can see, the particles first spread out to explore the function landscape and then move closer and closer to the global minimum further in the optimization. }
    \label{fig:pso-on-ackley-func}
\end{figure}

The behavior of the  PSO algorithm is sketched in Figure~\ref{fig:pso-on-ackley-func}, where PSO is used to optimize over the Ackley function used for benchmarking. We observe numerous local minima, in which the algorithm could get stuck, but it proceeds to the global minimum.

For the sake of completeness, it is to mention that there are a number of related  optimizers also based on interacting particle systems, which have gained lots of popularity in recent years as they yield stronger theoretical guarantees with respect to their convergence rate. In particular \textit{consensus-based optimization} models the movement of the particles using Brownian motion, which is in favour of the exploration capabilities of the particles \cite{totzeck2021trends, carrillo2021consensusbased, carrillo2018analytical, carrillo2020consensusbased, fornasier2021consensusbased-convergence, fornasier2021consensusbased-mean-field}. While motivated by implementation efficiency of the PSO algorithm we consider only the PSO method for the purpose of this paper, we find it a very interesting research direction to explore the potential of alternative optimization methods for our problem both in theory and practice.

\section{Self-Supervised Sparse Random Features for Surrogate Modelling}\label{sec:self-supervised-surrogate-modeling}

The method we propose in this paper combines the three aforementioned approaches to design and calibrate a surrogate model. The dimensionality is reduced via KPCA with parameters $\theta$ optimized via PSO and a surrogate model for the reduced dimension is designed as a sparse random feature expansion (LOS-RFE). Key to our approach is that these steps are not performed sequentially but the method is self-supervised -- the parameters are constantly updated in view of the success of the combined method. 

% In this section, we present a self-supervised surrogate modelling approach, which is key to this paper and allows to find optimal surrogate models via dimensionality reduction and sparse random feature expansion. 

The approach of enhancing a model's performance via self-supervision is especially common in the field of computer vision \cite{hur2020selfsupervised, gojcic2021weakly, mittal2020just} and has been just recently applied to surrogate modelling by Lataniotis et al., \cite{lataniotis2020extending}.
% Self-supervision allows to tune the parameters of one model with respect to the outputs of a second model. 
Such techniques are particularly useful when the available data is of insufficient resolution or in the case of scarce data, which is in the focus of this work. 
Further, a self-supervised approach eases the process of fitting a model in a data-driven setting as the incorporated algorithms feedback themselves.

To implement our self-supervised method, we split the experimental design $\X=\{\xb_1, \ldots, \xb_n\}$ and the  corresponding model response $\Y=\left\{\M(\xb_1), \ldots, \M(\xb_n) \right\}$ into two parts, a training set  $\X_{train}$ used to learn the coefficients of the random features and a validation set $\X_{val}$ used to supervise the learning process, i.e., to update to iteratively update the KPCA parameters. Note that that the validation set is different from the test set $\X_{test}$; the latter is never used in the learning process. That is, the data set is now divided into three parts of cardinalities $ N_{val}+N_{train}+N_{test}=n$.

 A challenge we are facing is that  it is not clear a priori which intrinsic dimension $k^{\star}$ to choose. Consequently, we explore multiple potential intrinsic dimensions in a candidate set $D \subset \N$. For each $k\in D$, we use PSO to simultaneously optimize the parameter $\theta$ for the KPCA dimensionality reduction map $\tau^k_{\theta}$ and the coefficients $c$ of the LOS-RFE surrogate model $f_c$ for the data set in the reduced dimension.

In the initialization step, $t=0$, of the PSO algorithm, we  randomly generate $P \in \N$ particles, $\{\theta_p \in \R^k, p=1,\ldots P\}$ representing the initial KPCA parameter vector $\theta$ and set $V_p(0)=0$. We then iterate over $t$ and repeat the following steps. 
%Each particle in the swarm of size $P \in \N$ represents  a vector of parameters $\theta_p$. 
For each $\theta_p$, the training set $\X_{train} \subset \R^d$ is reduced to ${\Zc_{\theta_p, train}=\tau^k_{\theta_p}(\X_{train}) \subset  \R^k}$ by the respectively parameterized KPCA  map $\tau^k_{\theta_p}$. For the reduced training set $\Zc_{\theta_p, train}$ and the corresponding model response $\Y_{train}$, we optimize the parameters $c_p=(c_{1,p}, \ldots, c_{R,p})$ in  a surrogate model $f_{c_p}$ of  the form 
\begin{equation}
    \label{eq:los-rfe-sm}
    f_{c_p}(\mathbf{z})= \sum_{j=1}^R c_{j,p} \, \phi(\mathbf{z}; \w_j), 
\end{equation}
via a LASSO-inspired method (see below for details). Here, the $\phi(\mathbf{z}; \w_j)$ 
are fixed realizations of random features with sparse Gaussian weights $\w_j\in \R^k$ as discussed above.

Now, again for each $p=1,\ldots, P$, the key paradigm of PSO is used to update the parameter $\theta_p$ on the reduced validation set $ \Zc_{\theta_p, val}=\tau^k_{\theta_p}(\X_{val}) \subset  \R^k$ based on the loss 
\begin{equation}\label{eq:emp:error-1}
\widehat{\epsilon}(\Zc_{\theta_p, val},  \Y_{val}, f_c)=\frac{  \sum_{i=1}^{ N_{val}} \big(y^{val}_i- f_{c}(\zb_i)\big)^2}{ \sum_{i=1}^{ N_{val}}\big(y^{val}_i- \mu_y\big)^2}. 
\end{equation}
associated with the surrogate model $f_{c_p}$ learned in the first step. 
That is, we compute a velocity $V_p(t+1)$ as  a combination of the previous velocity $V_p(t)$ with a cognitive and a social term, see Section~\ref{sec:pso} for details. We then use this velocity to update the KPCA parameter $\theta_p$ via 
\begin{equation}
    \widehat{\theta}_p(t+1)= \widehat{\theta}_p(t)+ V_p(t+1). 
\end{equation}
If the PSO algorithm converges (which we observed to be the case in most instances), we obtain a pair of KPCA parameters $\theta\in \R^k$ and LOS-RFE parameters $\mathbf{c}=(c_1, \ldots, c_R)$ for each $k \in D$.

It remains to describe how we compute random feature coefficients $\cb$. 
  
As the number of surrogate models to fit per dimension $k$ is equal to the product of the swarm size and the number of iterations (the optimization is run for),  solving for the optimal sparse coefficients $\mathbf{c}^{\star}$ via the LASSO, as proposed in Section~\ref{sec:srfe}, in each iteration is computationally quite demanding. To overcome this computational bottleneck, we observe that exactly solving the LASSO is of little value during the initial stage of the PSO algorithm, as first we need to find a
rough direction for the swarm of particles to be moved to.  To find  such a direction for the swarm $\{\theta_p \in \R^k, p=1,\ldots, P\}$ in a cheaper way,  we use for the initial PSO steps the following ridge formulation 
\begin{align}
    &\underset{\mathbf{c} \in \mathbb{R}^d}{\operatorname{arg min}} \ \lambda \| \mathbf{c} \|_2 + \| \mathbf{Ac} - \mathbf{y} \|^2_2 \label{eq:ridge-for-init-phase}  
\end{align}
 for finding the coefficients $c$ in the surrogate model $f_{c}$ instead of the LASSO.  
Such a ridge formulation can be solved more efficiently, decreasing the running time per optimization step. 
% Such trade-off  between sparsity of the coefficients $c$ and the seed of iterations is justified by the fact that the initial iterations of the PSO explore the function landscape, that is required to push the swarm into the direction towards the optimal $\theta^{\star}$.
Once the particles start to converge, i.e. the particle position's change is less than a certain threshold $\eta \in \mathbb{R}$  
\begin{equation}
    \Vert \theta_{\text{step}, p} - \theta_{\text{step} + 1, p}  \Vert_2 \leq \eta, \quad \forall p = 1, \ldots, P
\end{equation}
where $\theta_{\text{step},p}$ describes the position of  particle $p$ in the current iteration, the switch is made to the LASSO formulation 
\begin{align}\label{eq:lasso-2}
    &\underset{\mathbf{c} \in \mathbb{R}^d}{\operatorname{arg min}} \ \lambda \| \mathbf{c} \|_1 + \| \mathbf{Ac} - \mathbf{y} \|^2_2,
\end{align}
as described in Section~\ref{sec:srfe}, in order to find sparse random feature expansion for the computational model~$\M$.

At the end,  as soon as for  dimension $k \in D$ optimal parameters $\theta$ and $c$ have been found by the algorithm, we choose the composition of $\tau^{k^{\star}}_{\theta^{\star}}$ and $f_{c^{\star}}$, which enjoy the best performance among all $k\in D$, as the target surrogate model.

The approach just described is summarized in Algorithm~\ref{alg:self-supervised-sparse-random-feature-expansion-sparse-weights} with the following list parameters 
\begin{itemize}
    \item $P \in \mathbb{N}$ is the number of particles used for particle swarm optimization
    \item $p \in \mathbb{N}$ indicates particle index with $p \in \{1, 2, \dots, P\}$,
    \item $R \in \mathbb{N}$ is the number of random features in the LOS-SRF model $f_{c}$, 
    \item $D \subset \mathbb{N}$ is the set of potential intrinsic dimensions;  $D$ has to be either set by a user or has to be defined by a hyper-parameter search\footnote{A possible method to find a suitable set $D$ could be a coarse to fine approach. Starting with a few samples from a set with larger steps should yield single dimensions $d \in \mathbb{N}$ around which more dimensions $d^{\prime} = d \pm \{1, 2, \dots\}$ would be sampled to test.}
    \item $n_{\text{iterations}} \in \mathbb{N}$ indicates the number of iterations to optimize over the projection into a lower dimensional space of dimension $k$.
\end{itemize}
Moreover, for the purpose of this paper,  we use only different real-valued basis functions s.a. $\cos, \sin$ and $ReLU$, reducing the required memory drastically, as the targets are solely real-valued as well. 

\begin{algorithm}[ht]
    \caption[{Self-supervised Sparse Random Feature Expansion for Surrogate Modelling in High Dimensions}]{Self-supervised Sparse Random Feature Expansion for Surrogate Modelling in High Dimensions}
    \label{alg:self-supervised-sparse-random-feature-expansion-sparse-weights}
    \SetKwInOut{input}{input}
    \SetKwInOut{output}{output}    

    \input{ $ \{ (\mathbf{x}, y)_i \}^N_{i=1}$ where $\mathbf{x} \in \mathbb{R}^d$; $n,n_{\text{iterations}}, q,R,P \in \mathbb{N}$; $\eta \in \mathbb{R}$; $ D = \{k_1, \dots, k_m \vert k_j \in \mathbb{N}, k_j \ll d \} $ }

    % return optimal weights and min. error 
    \output{optimal params. for dim. reduction, surrogate modelling and final loss: $\theta^{\ast}, \mathbf{c}^{\ast}, \hat{\epsilon}^{\ast}$}
    
    \BlankLine  % empty line
    \texttt{ridge-reg} $\gets$ True

    \BlankLine  % empty line 
    ref. alg for KPCA and ref. alg. for PSO
    % init variables 
    \tcc{initialize parameters}
    $\theta_{0,p} \gets \mathcal{U}[0,1]^d, \text{for all } p \in \{1, 2, \dots, P\}$\;
    
    \BlankLine  % empty line

    \tcc{iterate over possible lower dimensions}
    \For{$k$ \KwTo $D$}{

        \tcc{optimize reduction and fit model in reduced dim.}
        \For{step \KwTo\Range{$n_{\text{iterations}}$, $\text{for all } p \in \{1, \dots, P\}$}}{
    
            \BlankLine  % empty line

            \tcc{reduce the data according to current parameters}
            $\mathbf{x}^{\prime}_{\text{train}}, \mathbf{x}^{\prime}_{\text{val}} \gets \tau^k_{\theta_{\text{step},p}}(\mathbf{x}_{\text{train}}, \mathbf{x}_{\text{val}})$, $\forall \mathbf{x}_{\text{train}} \in \mathcal{X}_{\text{train}}$ and $\forall \mathbf{x}_{\text{val}} \in \mathcal{X}_{\text{val}}$\;
            \BlankLine  % empty line
    
            \tcc{fit sparse random feature expansion acc. to Algorithm~\ref{alg:srfe}}
    
            \eIf { \texttt{ridge-reg} is True }{
                
               $f^{\ast} \gets \psi^q_{R, \text{Ridge}}(\mathbf{x}^{\prime}_{\text{train}}, \mathbf{y}_{\text{train}})$
                
            }{
                $f^{\ast} \gets \psi^q_{R, \text{LASSO}}(\mathbf{x}^{\prime}_{\text{train}}, \mathbf{y}_{\text{train}})$
            }
            
            $\mathbf{\hat{y}}_{\text{val}} \gets f^{\ast}(\mathbf{x}^{\prime}_{\text{val}})$\;
            \BlankLine  % empty line
    
            \tcc{compute loss between prediction from low dimensional data and ground truth}
            $\hat{\epsilon}_{\text{step}, p} \gets \mathcal{L}(\mathbf{y}_{\text{val}}, \hat{\mathbf{y}}_{\text{val}})$\;
            
            \BlankLine  % empty line
            
            \tcc{optimize reduction parameters, yielding config. for upcoming iteration}
            $\theta_{\text{step} + 1,p} \gets \mathbf{PSO}(\hat{\mathbf{\epsilon}}_{\text{step}}, \{\theta_{\text{step}, p}\}_{p=1}^P )$\;
            
            \BlankLine  % empty line        
            \If{ $ \Vert \theta_{\text{step},\cdot} - \theta_{\text{step} + 1,\cdot}  \Vert_2 \leq \eta$ }{
                \texttt{ridge-reg} $\gets$ False
            }
    
        } 
}

    \BlankLine  % empty line

\end{algorithm}

As a general note, the PSO method was chosen in the proposed setup as it only requires a notion of the current best particle in the swarm as well as the position for which the current particle obtained its best value. In contrast to that, popular gradient-based methods require the gradient of a function at a certain position and sometimes even the second derivative. While those methods can be very efficient for large problems, as e.g. with Neural Networks containing thousand of parameters, they are not used in the current setting as they would interfere the modularity of the approach\footnote{While presenting a specific instance of a self-supervised algorithm, the aim should be to ease changes in the algorithms used for dimensionality reduction and surrogate modelling.}. Further, PSO is well suited for non-convex optimization problems as the interplay between the particles helps to prevent the optimizer getting stuck in small local minima, without having to adapt the update parameters during optimization, as it is for example done with learning rate schedulers for gradient descent. Thus, the choice of a metaheuristic for optimization is in alignment with the data-driven mentality of tackling surrogate modelling.

We note that our method is designed for a medium-sized number of observations. For a larger number $N$ of observations we  advise to use different dimensionality reduction method due to the complexity of $\mathcal{O}(N^2)$ arising in KPCA when computing the kernel matrix. Due to the modularity of our approach this can be done without interfering with other parts of the algorithm.

Finally, the remaining hyper-parameters required in Algorithm~\ref{alg:self-supervised-sparse-random-feature-expansion-sparse-weights}, such as $R,q$, can be found via hyper-parameter search. Common methods for hyper-parameter search include simple (but inefficient) grid-search, Bayesian search or random search \cite{bergstra2011algorithms, bergstra2012random, liashchynskyi2019grid, wu2019hyperparameter}. In the experiments of Section~\ref{sec:numerical-experiments} we use Bayesian search.

\newpage 
\section{Numerical Experiments}\label{sec:numerical-experiments}

In this section,  we test the proposed surrogate modelling approach based on LOS-RFE numerically and compare the results to state-of-the-art surrogate modelling techniques based on neural networks and polynomial chaos expansions.  

In our numerical tests, we use both realistic and synthetic data enabling a thorough analysis of the methods with regards to practical applicability as well as their ability of correctly conceiving the data structure.
The first dataset is generated from a function given in a closed form, which allows to compare the output of the surrogate model with the ground-truth data generated by this function.  The second dataset has been provided by the Department of Vehicle Safety at BMW Group and consists of  crash test simulations for real world settings.  
Both of the presented datasets have unstructured inputs and scalar outputs. While the intrinsic dimension is approximately known for the first dataset, it is unknown for the second one.

In the following,  the three different methods are compared against each other with respect to the empirical generalisation error \eqref{eq:emp:error}, both on training and test data. We start by describing the data sets and also the methods in some more detail.

%-----------------------------------------------------------------------------
\subsection{Description of Datasets}

{ \textbf{ First Dataset}.}  
The Sobol function, often also called the G-function \cite{simulationlib, saltelli2008, lataniotis2020extending}, is common for benchmarking in the context of UQ. As the properties of the function, such as the shape or approximate intrinsic dimension, are known and can be configured via an appropriate choice of parameters, the function yields a rich class of test cases. 
%with arbitrary high dimensions. 
In general, for some $K \in \mathbb{N}$,  the Sobol function is defined as 
\begin{equation} \label{eq:sobol-function}
    f_{\text{Sobol}}(\xcb) = \prod^K_{i=1} \frac{|4\xcb_i - 2| + u_i}{1 + u_i},\ u_i \in \mathbb{R}_{+}
\end{equation}
where $\xcb = \{\xcb_1, \dots, \xcb_K\}$ are independent random input variables drawn from the uniform distribution $\mathcal{U}[0,1]$, which in the setting of UQ hold the uncertainty of the system. The vector of parameters ${\mathbf{u} = \{u_1, \ldots, u_k\} \in \mathbb{R}^K_+}$ controls the intrinsic dimension as it weights $\xcb_j$'s effect on the output value of $f_{Sobol}$. 
To be more precise, the effect of each input variable $\xcb_j$ to the output $f_{Sobol}(\xcb)$ is inversely
proportional to the value of $u_i$:  relatively small values of $u_i$ result in a relative importance of $\xcb_i$ for $f_{Sobol}(\xcb)$. 
Finally, the parameter $K$ defines the input dimension and allows to scale to extremely large dimensions. 

For our first dataset, we aim to choose the vector $\mathbf{u}$ to yield an intrinsic dimension of about $6$ with an input dimension of $K=20$. For this purpose, following  \cite{konakli2016global, kersaudy2015new, lataniotis2020extending}, the vector $\mathbf{u}$ is set to
\begin{equation} \label{eq:sobol-func-constants}
    \mathbf{u} = \{1, 2, 5, 20, 50, 100, 500, \dots, 500\}. 
\end{equation}
Thus, the first 6 variables can provide a compressed representation of the data, while a few extra dimensions might be valuable to capture the information provided by dimensions 7-20. Picking the exact number of intrinsic dimensions as target dimension would be called expert's knowledge. However, such information is usually not available, which is why  we explore multiple  potential intrinsic dimensions.

In our experiments, the input data $\xcb_i \in \mathbb{R}^d$ for  \eqref{eq:sobol-function} were sampled from $[0,1]^d$ using the Sobol sampling schema \cite{sobolsampling}, allowing for a better coverage of the domain compared to pure pseudo-random number generators.

\begin{figure}
    \centering
    \includegraphics[width=.6\linewidth]{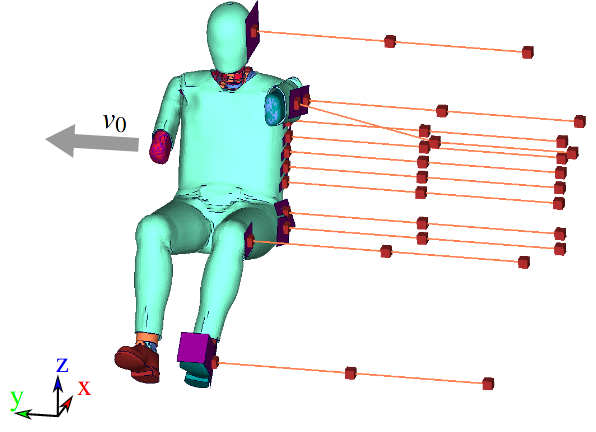}
    \caption[Snapshot of the data-generating crash simulation]{Snapshot of the data-generating simulation showing the full-scale dummy experiencing a lateral force in its initial position. Image taken from \cite{Jehle2021, jonasjehlephd}.}
    \label{fig:crash-test-dummy}
\end{figure}

{ \textbf{  Second Dataset.}}
Here we use a dataset provided by the Department of Passive Safety at the BMW Group, which consists of crash test data obtained from computational simulations of a highly accurate Finite-Element (FE) model. As the evaluation of the FE model tends to run for several hours for one setting of input parameters, UQ directly on the data-generating model is infeasible. Therefore, the provided dataset places us in a realistic setting where a highly accurate but cheap to evaluate surrogate model is needed to approximate the original FE simulation.

Each BMW FE simulation is configured by a uniformly distributed random vector $\xcb_i \in \mathbb{R}^d$, for $d=36$ and  $i=1, \dots, 600$, where the different components of  the $\xcb_i$  correspond to certain attributes such as car speed, impact angle etc.  While the precise nature of this distribution cannot be revealed in this work due to intellectual property 
constraints \cite{jonasjehlephd}, we stress that we have used the same data for all three methods.  The simulations give rise to target values (data) $\Y$ describing a head-injury value in the setting illustrated in Figure~\ref{fig:crash-test-dummy}. In contrast to the first dataset, increasing the number of observations is a bottleneck as it requires highly expensive computations.

\subsection{Description of Surrogate Modelling Techniques Used for Comparison}\label{sec:compared-models}

The first method chosen for comparison, described by Lataniotis et al. \cite{lataniotis2020extending}, is self-supervised similar to the method proposed in this paper. As a key building block in addition to dimensionality reduction, it uses a polynomial chaos expansion. More precisely, a  genetic algorithm is used to find a suitable set of parameters for dimensionality reduction, and then the reduced data is iteratively approximated by a polynomial chaos expansion.

As the second method, we chose neural networks as comparison given that they inherently incorporate dimensionality reduction if one reduces the number of nodes layer by layer. Further, they have the ability to model complex relationships due to the nonlinear activation in each node. Neural networks have proven themselves
to be the model of choice in several disciplines \cite{nnaer2019, nnsmpfrommer2018optimisation, zhang2021, papadopoulos2018, nnsmtripathy2018deep}. However, they tend to require a lot of data to optimize their parameters. In the setting of Uncertainty Quantification, it is usually the case that there are only a limited number of samples available for fitting a model, so the data scarce setting deserves particular attention.

The PCE-based approach was fitted according to the description given by Lataniotis et al. \cite{lataniotis2020extending}, i.e. iterating over the set of possible latent dimensions and optimizing the dimensionality reduction's parameters with respect to  the surrogate's performance in that dimension. The neural networks were constructed by using fully connected layers with the ReLU activation function. The number of nodes per layer was chosen to minimize the training error using cross-validation as well as minimizing the generalisation gap, i.e. the difference between errors on training and validation data. The final network for the first dataset, for example, had the following number of nodes per hidden layer: $20 \to 10 \to 8 \to 4 \to 1$.

\subsection{Numerical Comparison}\label{sec:numerical-results-comparison}

\subsubsection{Performance on the Sobol Dataset}\label{subsec:numerical-results-sobol}
% {\mr use $10^5$ for val. in acc. to Lat. }
To test different surrogate models  from the theoretical prospective, we use the Sobol function $f_{\text{Sobol}}$ with parameters $\mathbf{u}$ as in \eqref{eq:sobol-func-constants} to generate the dataset consisting of $N_{\text{train}} = 800$ samples for training, $N_{\text{val}}=1200$  for validation, and another $N_{\text{test}}=2000$ for testing. 
Such a division of the training and test data was motivated by a more detailed error analysis of the models' ability for generalization to unseen data points. 

\begin{figure}[ht]
    \centering 
    % ------------------------------------------------------------------------
    \includegraphics[width=.7\linewidth]{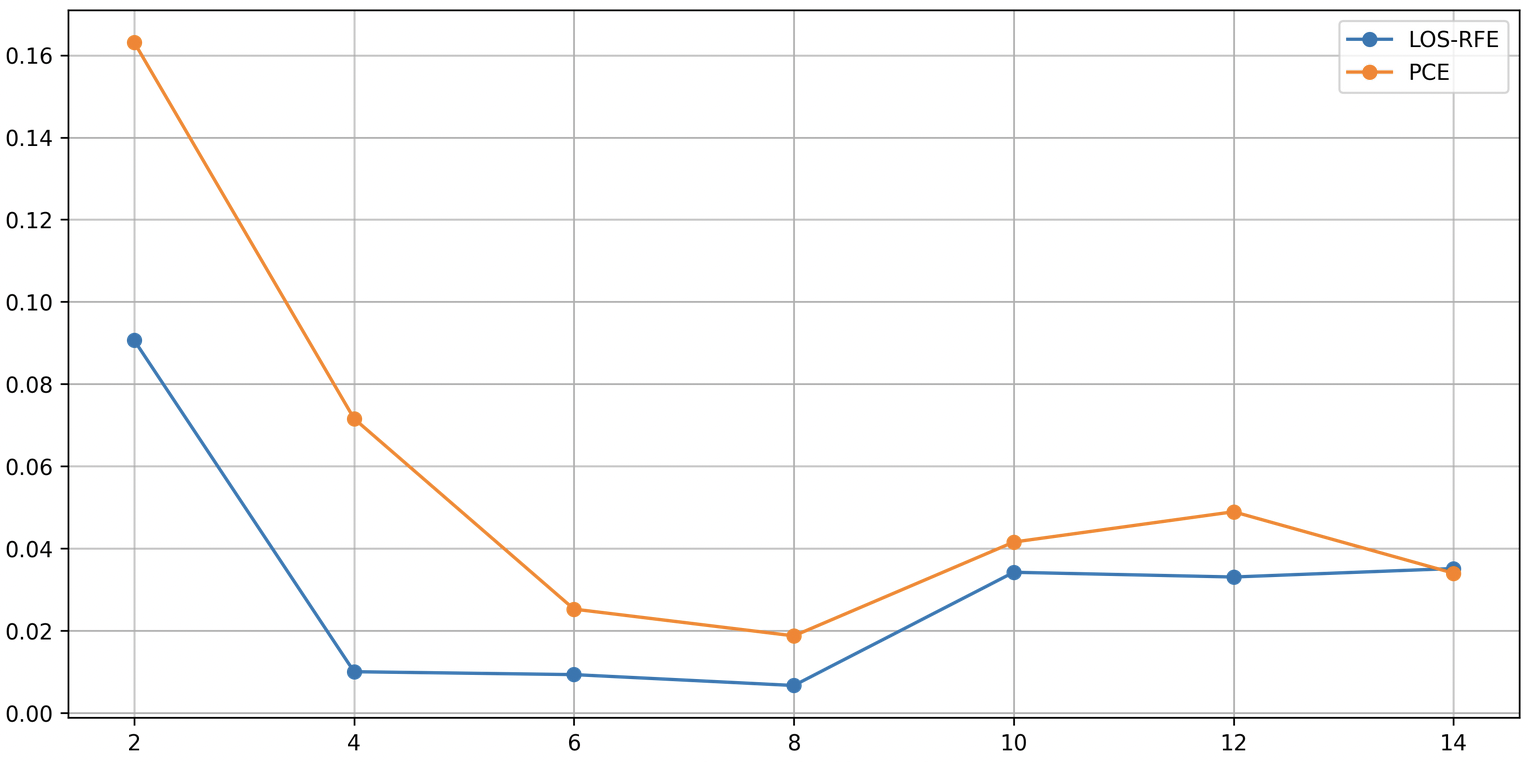}
    % ------------------------------------------------------------------------    
    \caption[Performance history of PCE-based and LOS-RFE on Sobol data]{ 
    % { \mr adapt the opt. dimensions }
    The figure shows the performance history of the PCE-based approach and the proposed LOS-RFE approach on the Sobol data per latent dimensions in $D$ (x-axis) in terms of validation error (y-axis) in the final stage of  the self-supervised learning process. 
    As one can observe for latent dimensions $k \in \{6,8\}$, the LOS-RFE approach and  PCE-based approach shows similar performance, but LOS-RFE approach enjoys lower validation error than the PCE-based one.  The neural network approach is not included here as it does not have an explicit dimensionality reduction step.    }
    \label{fig:performance-hist-sobol}
\end{figure}

\begin{figure}[ht]
    \centering    
    % PCE-based results -------------------------------------------------------
    \subfloat[PCE-based, $\hat{\epsilon}_{\text{train}} = 0.0115$ \label{fig:sobol-data-full-PCE-based-dim6-dim8-train}]{
        \includegraphics[width=.3\linewidth]{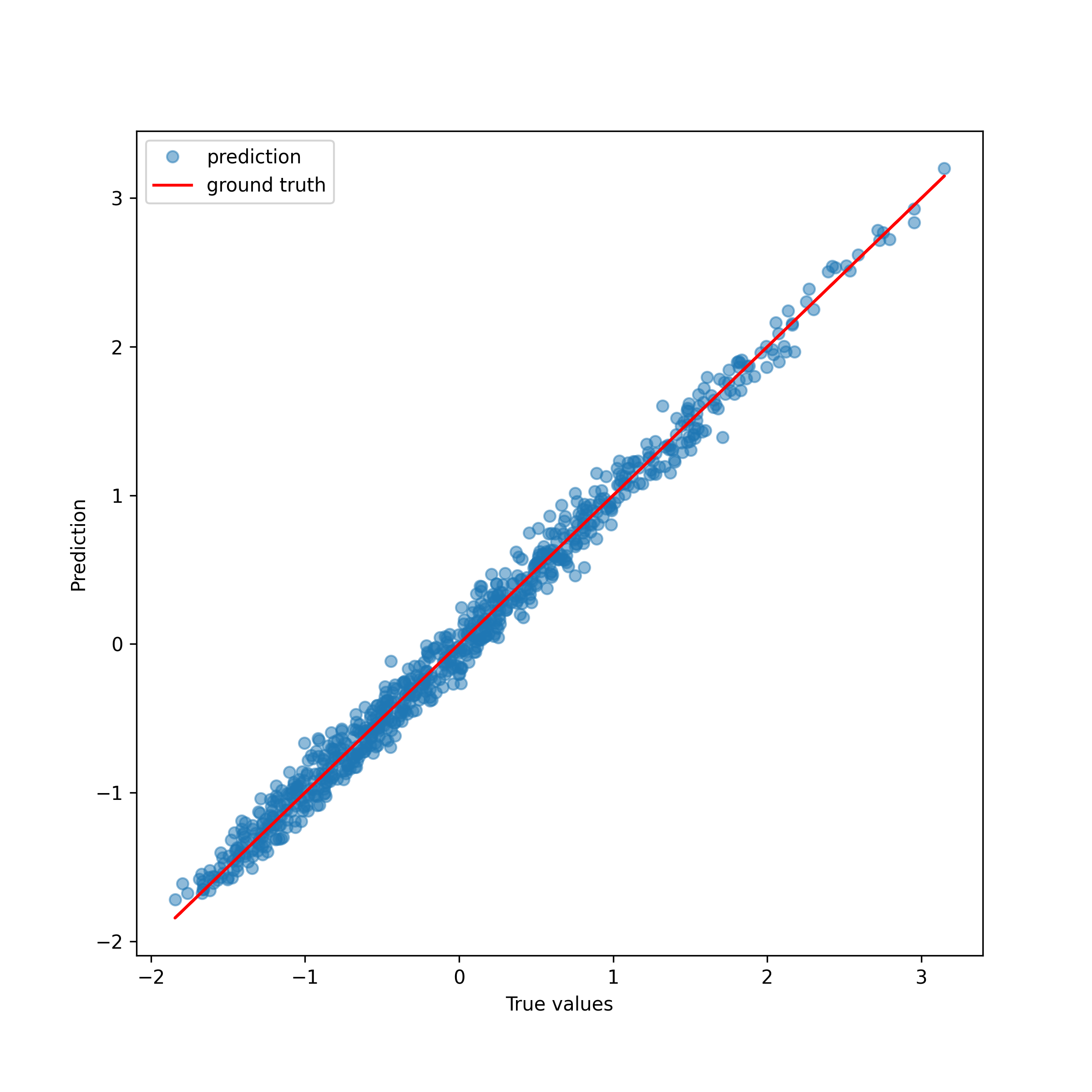}
    }
    % neural net ------------------------------------------------------------------
    \subfloat[NN, $\hat{\epsilon}_{\text{train}} = 0.0048$ \label{fig:sobol-data-full-train-nn}]{
        \includegraphics[width=.3\linewidth]{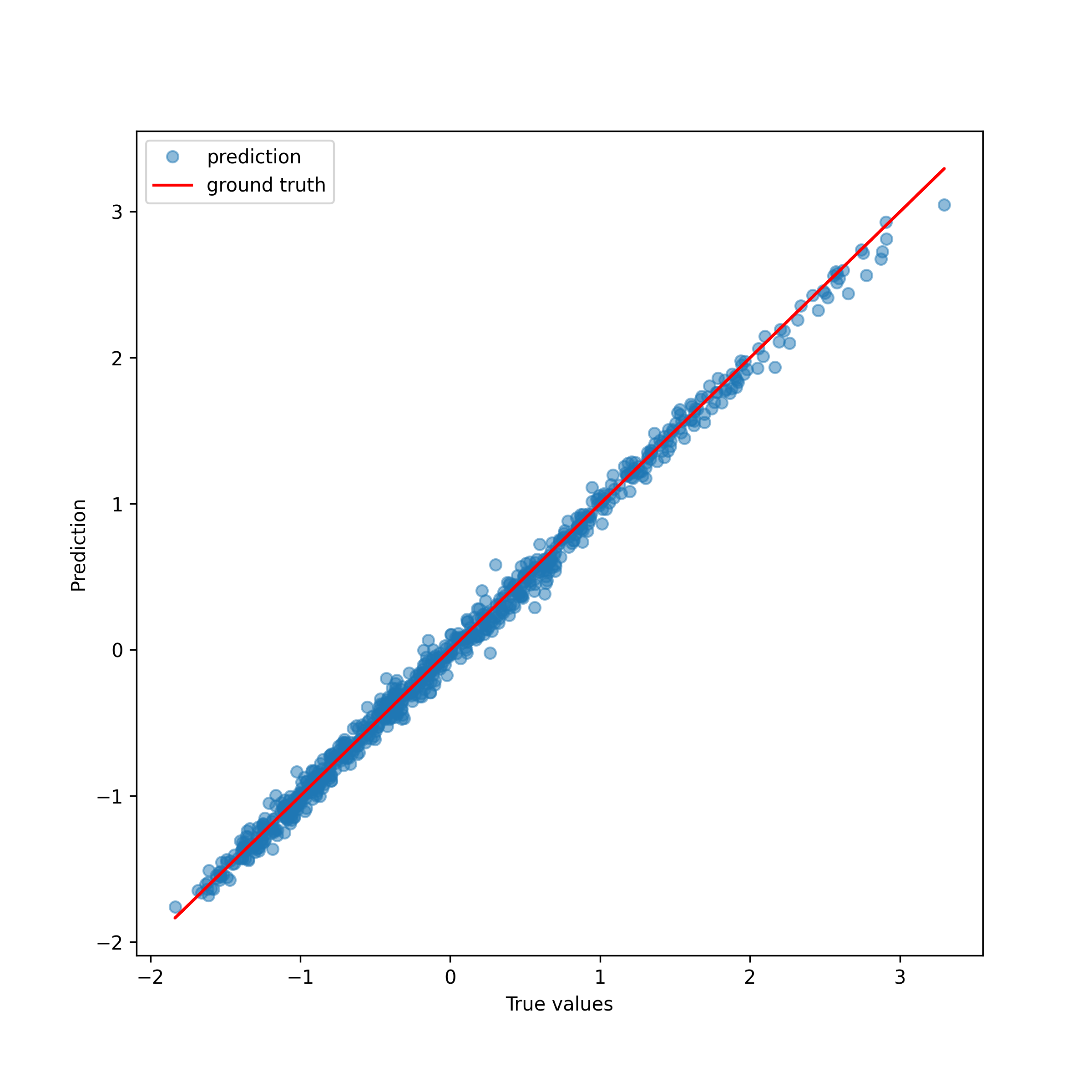}
}
    % ours ------------------------------------------------------------------------
    \subfloat[LOS-RFE, $\hat{\epsilon}_{\text{train}} = 0.0032$ \label{fig:sobol-data-full-train-srfe}]{
        \includegraphics[width=.3\linewidth]{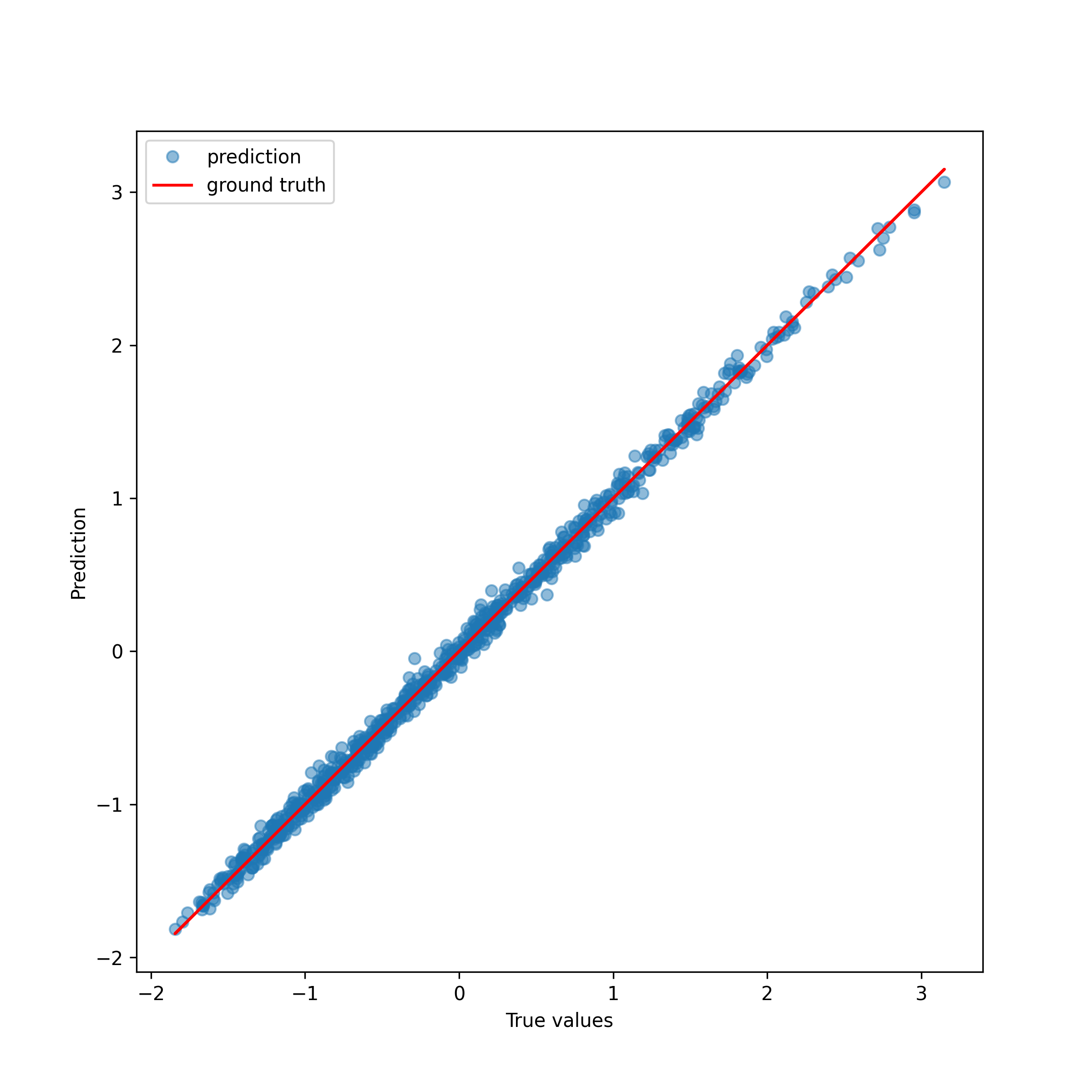}} 
    
    \vspace{0.2cm}
    
    \subfloat[PCE-based, $\hat{\epsilon}_{\text{test}} = 0.0171$ \label{fig:sobol-data-full-PCE-based-dim6-dim8}]{
        \includegraphics[width=.3\linewidth]{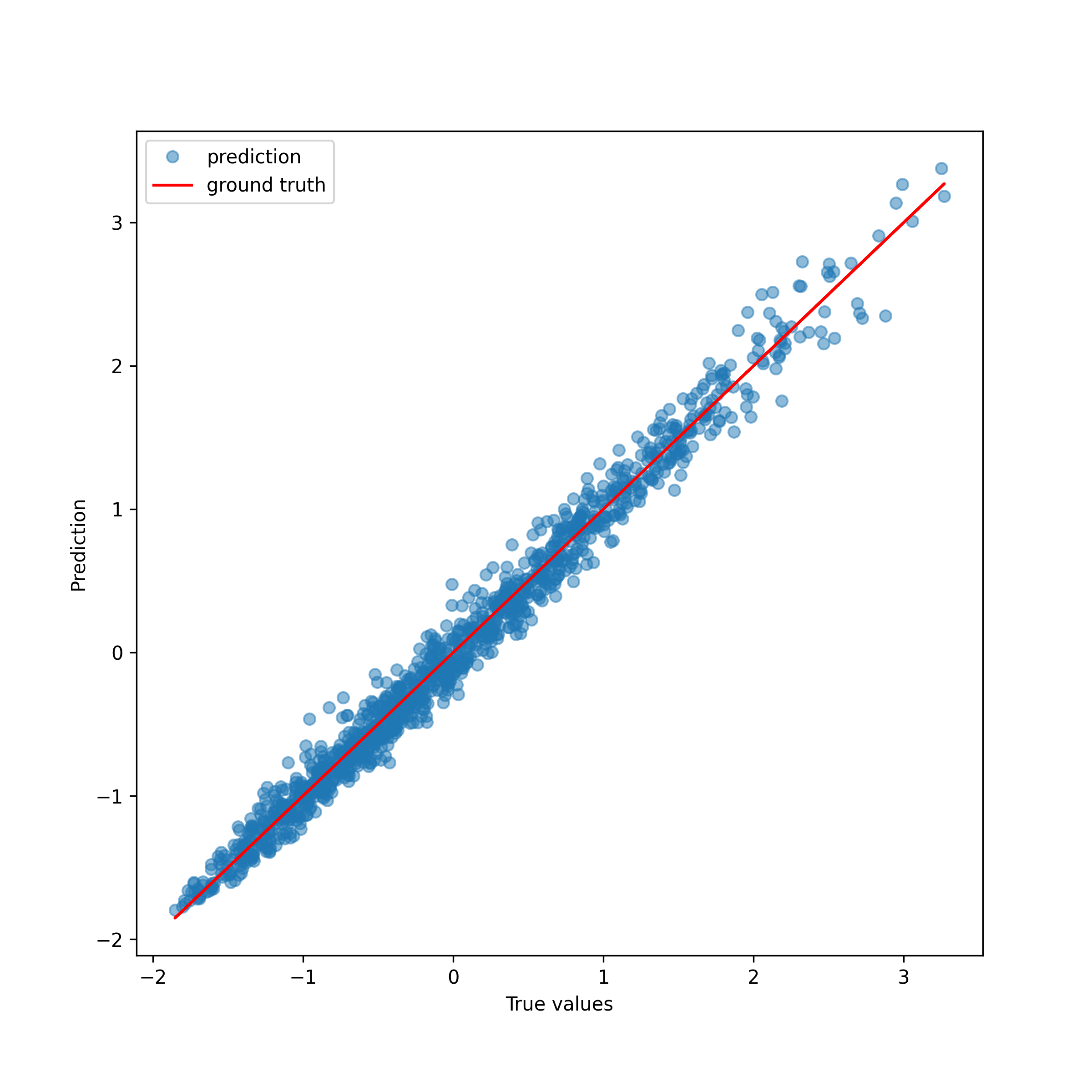}
    } 
    % -----------------------------------------------------------------------------
    \subfloat[NN, $\hat{\epsilon}_{\text{test}} = 0.0151$ \label{fig:sobol-data-full-val-nn}]{
        \includegraphics[width=.3\linewidth]{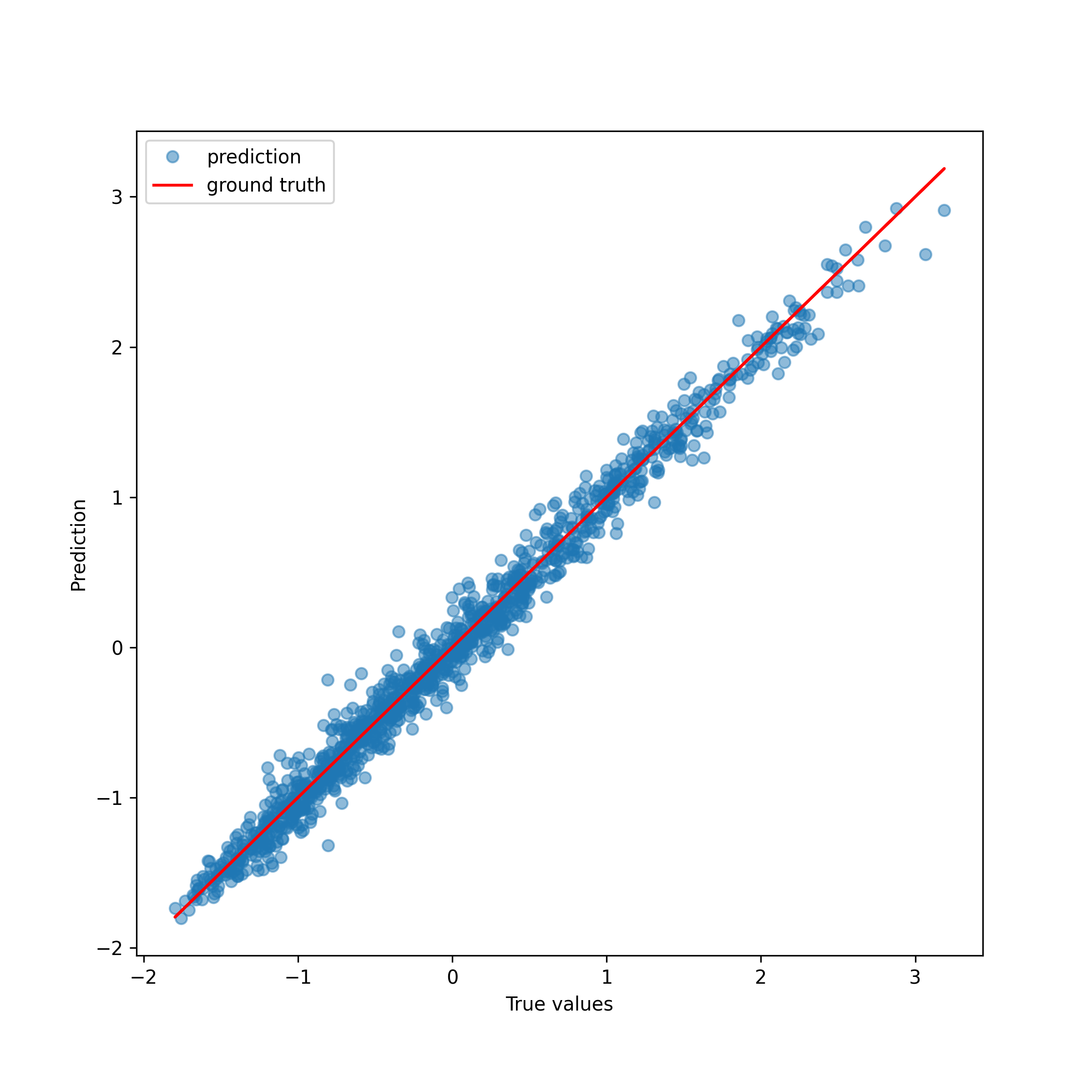}
    }  
    % -----------------------------------------------------------------------------
    \subfloat[LOS-RFE, $\hat{\epsilon}_{\text{test}} = 0.0062$ \label{fig:sobol-data-full-val-srfe}]{
        \includegraphics[width=.3\linewidth]{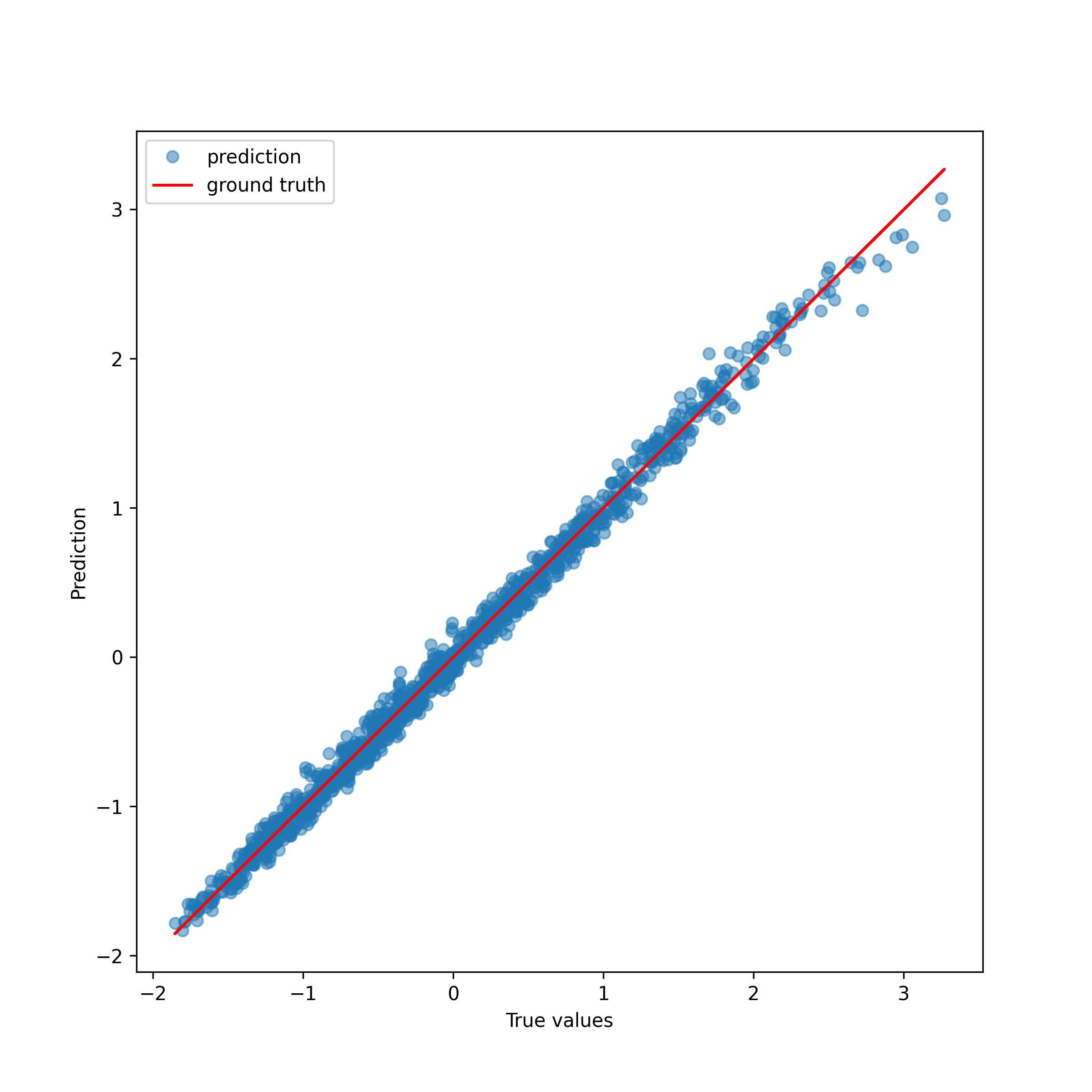}
    }
    
    \caption{The performance of different surrogate modelling approaches on the training ($N_{\text{train}}=800$ and $N_{\text{val}}=1200$) as well as the test data ($N_{\text{test}}=2000$) generated by the Sobol function parameterized by $\mathbf{u}$ as in \eqref{eq:sobol-func-constants}.
    % , chosen when generating the data. 
    Figure \ref{fig:sobol-data-full-PCE-based-dim6-dim8-train}-\ref{fig:sobol-data-full-PCE-based-dim6-dim8} correspond to the PCE-based approach \cite{lataniotis2020extending} with a best performing latent dimension of $k=8$, figure \ref{fig:sobol-data-full-train-nn}-\ref{fig:sobol-data-full-val-nn} depicts results for a neural network, and figure \ref{fig:sobol-data-full-train-srfe}-\ref{fig:sobol-data-full-val-srfe} shows the performance of the LSO-RFE for $k=8$ as well. 
    }  
    \label{fig:performances-on-sobol}
\end{figure}

For these data sets, firstly,  the above-described PCE-based surrogate model was trained and obtained according to \cite{lataniotis2020extending} for the set of potential intrinsic dimensions ${D=\{2,4,6,8,10,12,14\}}$. For each latent dimension $k\in D$, the data was projected into the lower $k$-dimensional space, using Kernel PCA with an anisotropic Gaussian kernel with parameters learned in a self-supervised process with respect to the model's performance during validation.  The highest  order of multivariate polynomials was chosen to be three as this showed to be the best trade-off between model complexity and ability to generalize in the given setting, see \cite{lataniotis2020extending} for more details. 
The method's performances over all intrinsic dimensions at the final stage of training process is illustrated in Figure~\ref{fig:performance-hist-sobol} with the best performing dimensions $k=8$. 
Such combination of dimensionality reduction and PCE, results in the empirical generalization errors \eqref{eq:emp:error}, for the best-performing dimensions $\hat{\epsilon}_{d=8} = 0.0171$\footnote{While Lataniotis \cite{lataniotis2020extending} reports a performance of $\hat{\epsilon} \approx 0.008$, it was not possible to reproduce those results. Also, the narrative of this subsection does not change with respect to  the optimal performance of the PCE-based method. Taking into account the optimal performance from the reference, the neural network would then yield the lowest performance, and the LOS-RFE method would still perform best, as shown in Table~\ref{tab:table-of-results}
.}. A detailed view on the predictions vs. ground truth values for the best performing dimension $k=8$ on the validation set and the test set is depicted in 
Figure~\ref{fig:sobol-data-full-PCE-based-dim6-dim8-train} and Figure~\ref{fig:sobol-data-full-PCE-based-dim6-dim8}, respectively. As one can see, the PCE-based surrogate performs almost in the same manner for on both sets, which means that the model has learn fairly well the underlying model~$f_{\text{Sobol}}$. 
%-----------------------------------------------------------------------------
% hist of performances for pce and ours 
% -----------------------------------------------------------------------------
% -----------------------------------------------------------------------------

% -----------------------------------------------------------------------------
% -----------------------------------------------------------------------------

Secondly, a neural network is fit on the same training data and evaluated on the same hold-out 
% validation 
test data. The neural network's performance on the training data and in the test data is shown in Figure~\ref{fig:sobol-data-full-train-nn} and Figure~\ref{fig:sobol-data-full-val-nn}, respectively. As an advantage, the model does not need a self-supervised projection to yield a suitable representation of the data as it is capable of learning this mapping within its layers. The neural network shows a superiority over the PCE-based approach as the network's error is roughly half of the PCE-based's error.

In order to generate the LOS-RFE-based surrogate model, we choose sparsity order $q=2$ and generate $R=10000$ random $\cos$-features as described in Section~\ref{sec:self-supervised-surrogate-modeling}. Although here we have even lower order of iterations between variables than for PCE, random features allow to generate a richer family of functions for surrogate modelling. This is also reflected in  numerical experiments leading to a better model fit on average, see  Figure~\ref{fig:performance-hist-sobol}.
For the LSO-RFE model as for the PCE-based one, the latent dimension $k=8$ was found to be optimal in the Sobol setting. However, the LSO-RFE method in combination with dimensionality reduction yields far lower errors than the PCE-based model as well as the neural network, both on validation and test data see  Figure~\ref{fig:sobol-data-full-train-srfe} and Figure~\ref{fig:sobol-data-full-val-srfe}. This indicates that the LOS-RFE model was able to generalize to the unseen data without overfitting, leading to test error $\hat{\epsilon}= 0.0062$. 

\begin{figure}[ht]
    \centering
    % NN ------------------------------------------------------------------------
    \subfloat[NN larger data, $\hat{\epsilon}_{
    \text{train}} = 0.0025$ \label{fig:sobol-data-large-nn}]{
        \includegraphics[width=.3\linewidth]{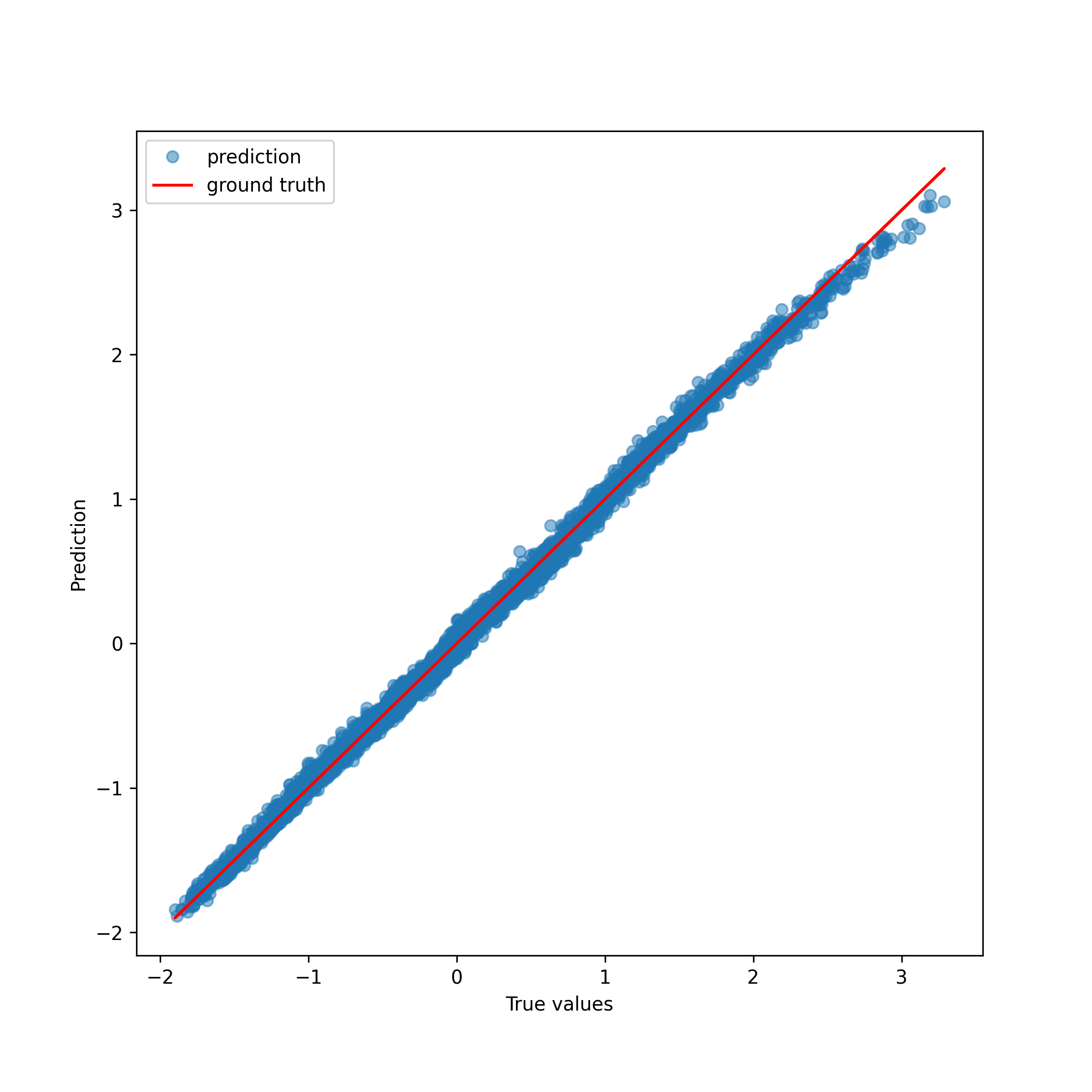}} \hspace{0.1\textwidth}
     \subfloat[LOS-RFE larger data, ${\hat{\epsilon}_{
    \text{train}} = 0.0041}$]{
        \includegraphics[width=.3\linewidth]{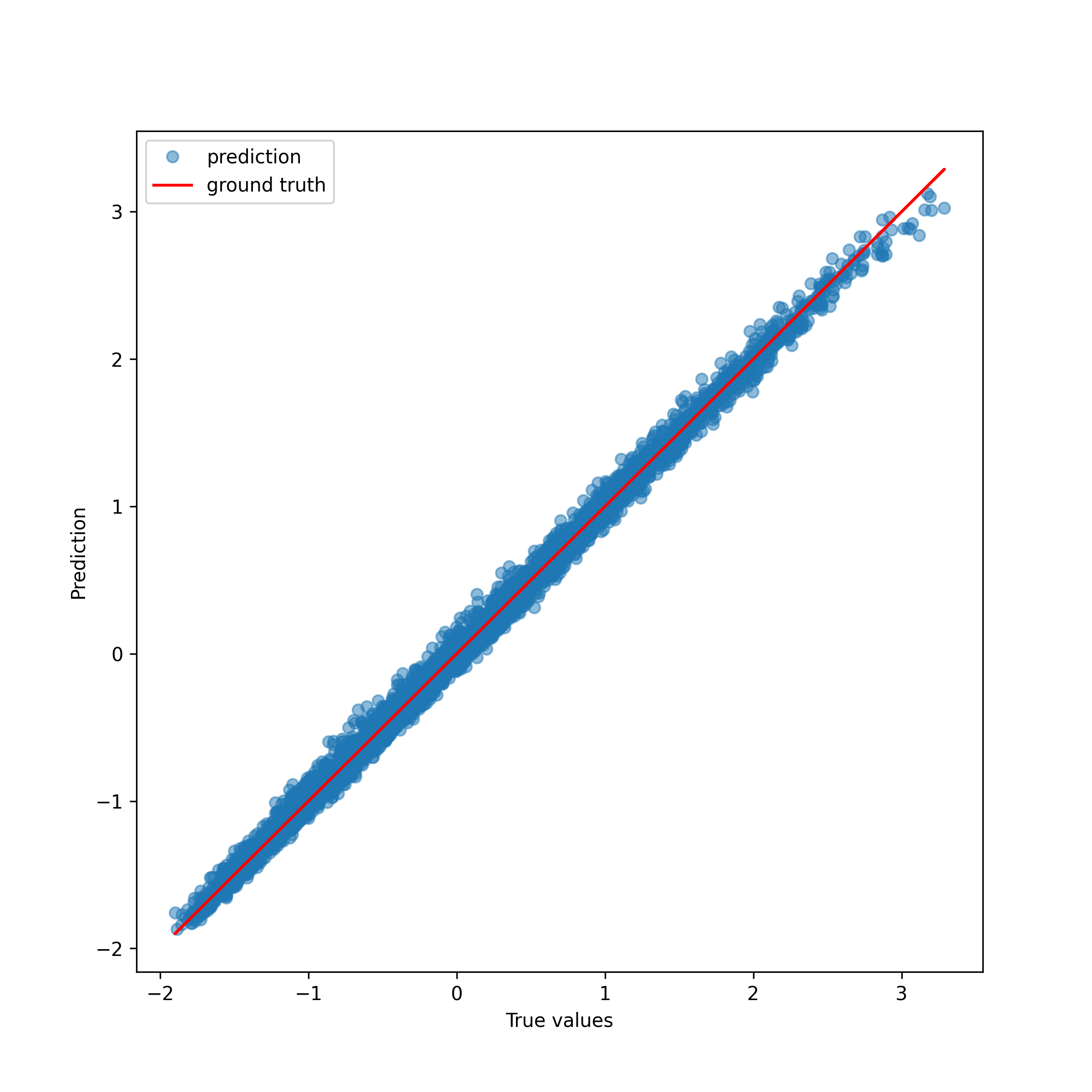}} 
   \vspace{0.2cm}
    % -----------------------------------------------------------------------
    % ours ------------------------------------------------------------------
    \subfloat[NN larger data, $\hat{\epsilon}_{
    \text{test}} = 0.0034$ \label{fig:sobol-data-arge-nn}]{
        \includegraphics[width=.3\linewidth]{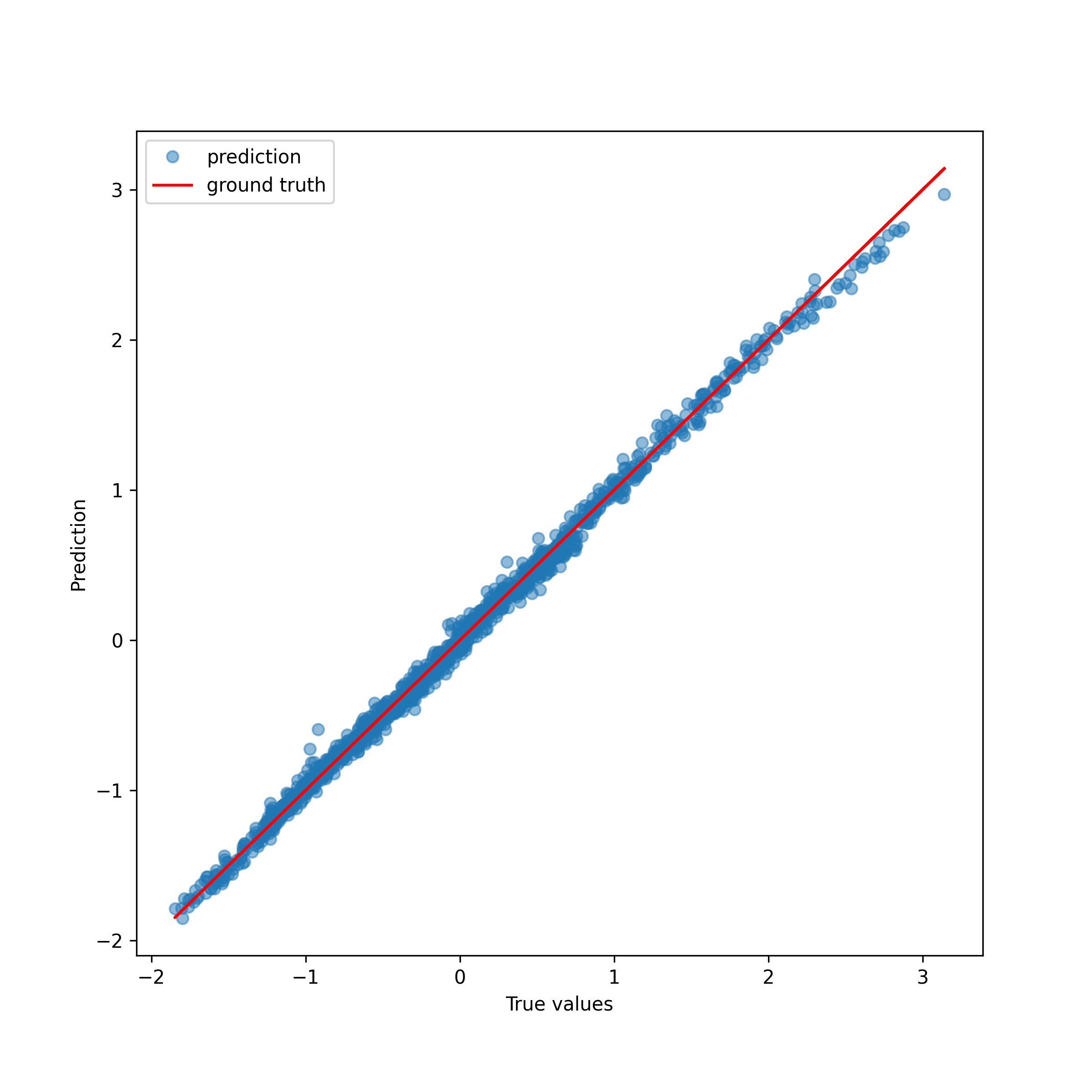}} \hspace{0.1\textwidth}
    \subfloat[LOS-RFE larger data, ${\hat{\epsilon}_{
    \text{test}} = 0.0052}$]{
        \includegraphics[width=.3\linewidth]{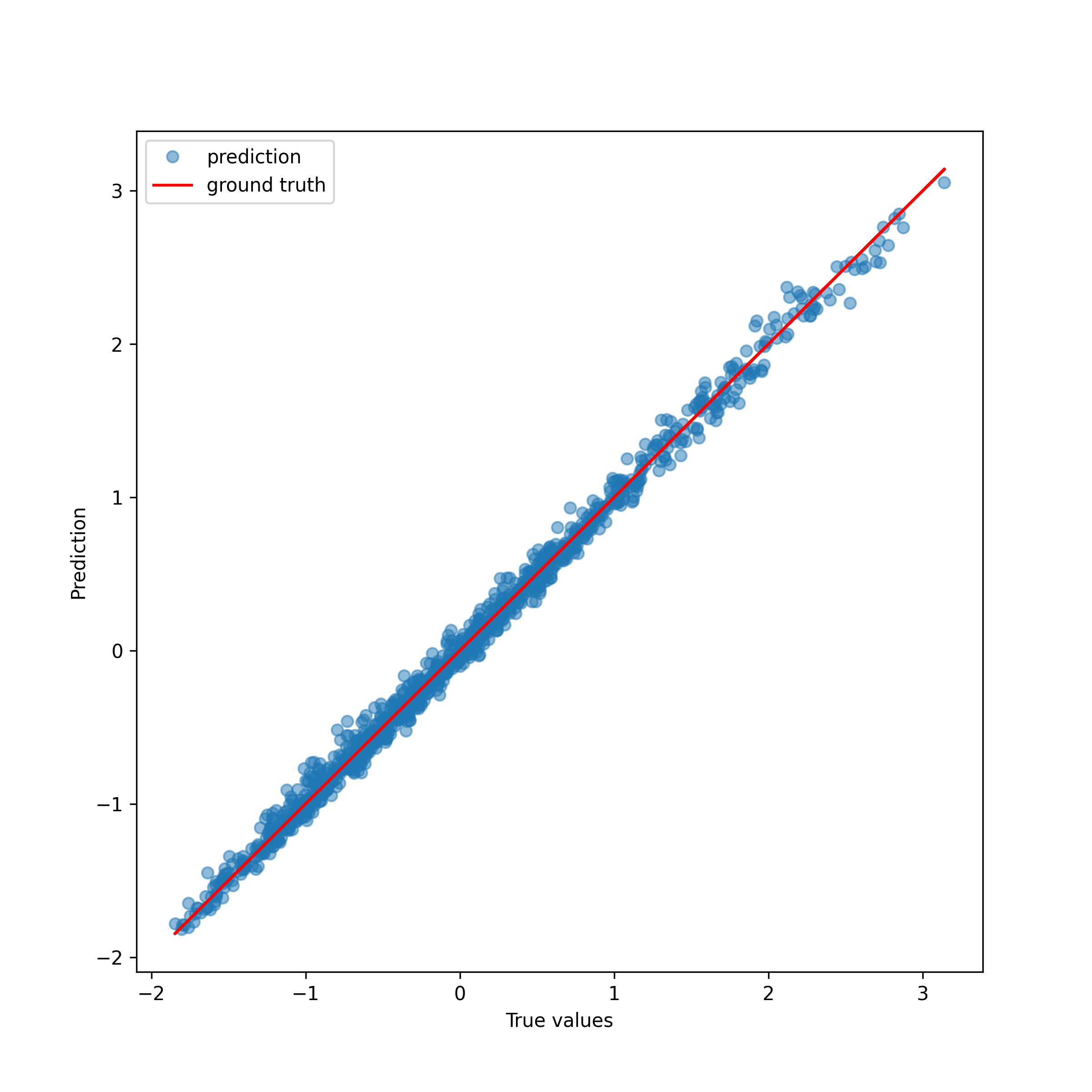}}
    % -----------------------------------------------------------------------------
    \caption[Performance of the proposed approach on the reduced Sobol dataset]{
    % {\mr change order of figures...  }
       The performance of the neural network and the LOS-RFE on an increased training dataset $N^{\ast}_{\text{train}}=4800$ compared to  $N_{\text{train}}=800$ as used to train the models shown in Figure~\ref{fig:performances-on-sobol}.
        % Figure~\ref{fig:sobol-data-full-PCE-based-dim6-dim8} and Figure~\ref{fig:sobol-data-full-srfe}. 
        The neural network approach has a clear advantage in this large data regime. The PCE-based approach is not included here due to computational limitations airing from its longer run times.
        }
    \label{fig:ours-and-nn-on-large-data}
\end{figure}

Not surprisingly, in the large data setting neural networks have a clear advantage over the two other methods. For example, if the data is enlarged to $N_{\text{train}}=4800$, a neural network performs better the LOS-RFE, as it is shown on Figure~\ref{fig:ours-and-nn-on-large-data}. This confirms that our method is particularly competitive in the scarce data regime, which often occurs in the surrogate modelling setting. 

% {\mr Third, the LOS-RFE  description: $R=10000 , \lambda=10^{-5}, q=2, $ }

\begin{figure}[ht]
    \centering
    % pce ------------------------------------------------------------------------
    \includegraphics[width=.7\linewidth]{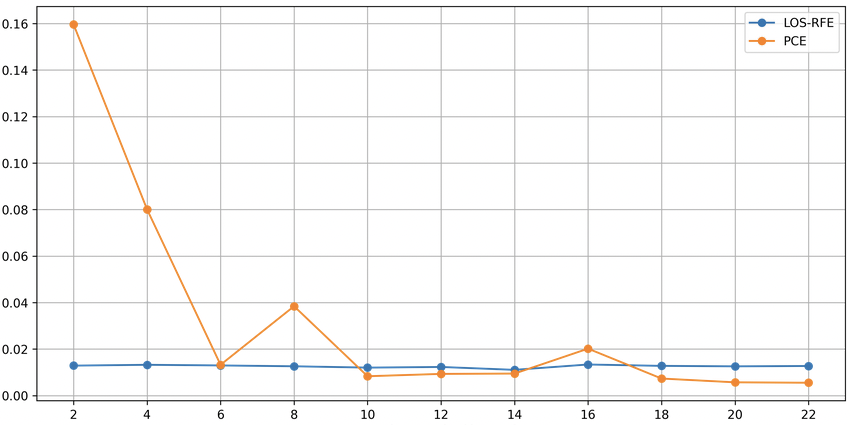}
    % ----------------------------------------------------------------------------
    \caption[Performance history of PCE-based and LOS-RFE on Crash data]{
        Similar to Figure~\ref{fig:performance-hist-sobol}, the figure shows the performances for the PCE-based method and the LOS-RFE
        % for different latent dimensions $k \in \{2, 4, 6, \dots, 20\}$ 
        on the crash data per latent dimensions in $D$ (x-axis) in terms of validation error (y-axis) in the final stage of  the self-supervised learning process. 
        % the final stage of training process. 
Regarding the PCE-based method, there is no a clear minimum but a trend towards higher dimensions. The ragged graph hints an unfinished optimization for $k=16$. Treating the observation at $k=16$ as outlier due to an unfinished optimization, the latent dimensions of $k=22$ seems to provide a good representation of the data.
        Similar to the PCE-based approach, the favorable dimensions for our approach are larger than for the Sobol data. In this case, the lowest validation error for LOS-RFE was attained for $k=14$.}
    \label{fig:performance-hist-pce-and-ours-on-crash-data}
\end{figure}
% -----------------------------------------------------------------------------
% -----------------------------------------------------------------------------

% { \mr in turn, if the data is enlarged to ..., the neural networks have a clear advantage over all methods. ... losses etc. }

%-----------------------------------------------------------------------------
% -----------------------------------------------------------------------------

\subsubsection{Performance on the Crash Test Dataset}\label{subsec:numerical-results-crash}

As mentioned above, the crash test dataset resulted from $N_{\text{total}}=600$ executions of a crash test FE simulation \cite{Jehle2021, jonasjehlephd}. The simulation itself was mainly focused on the mannequin inside the car rather than the deformations happening to the chassis. Due to the emergence of the dataset, the latent dimension is not known - differently from the Sobol dataset. 

For the comparison, the same three types of  models as for the Sobol data are  determined  on the new data sets with $N_{\text{train}} = 150$  elements for training and $N_{\text{val}} = 150$ for validation, $N_{\text{test}} = 300$ for testing, allowing to mimic a data-scarce scenario while holding enough observations back, to get a good error estimate.

\begin{figure}[ht]
    \centering
    % \subfloat[PCE-based, $\hat{\epsilon}_{\text{train}}=6.7397 \cdot 10^{-28}$]{
    %     \includegraphics[width=0.3\linewidth]{figures/pce_crash_training_pred.png}
    % }
    % \subfloat[NN-based, $\hat{\epsilon}_{\text{train}}=$ TODO]{
    %     \includegraphics[width=0.3\linewidth]{figures/nn_crash_train_pred.png}
    % }
    % \subfloat[PCE-based, $\hat{\epsilon}_{\text{train}}=$ TODO]{
    %     \includegraphics[width=0.3\linewidth]{figures/placeholder.png}
    % }
    % PCE-based results -------------------------------------------------------
    \subfloat[PCE-based, $\hat{\epsilon}_{\text{train}} = 6.7397 \cdot 10^{-28}$]{
        \includegraphics[width=.3\linewidth]{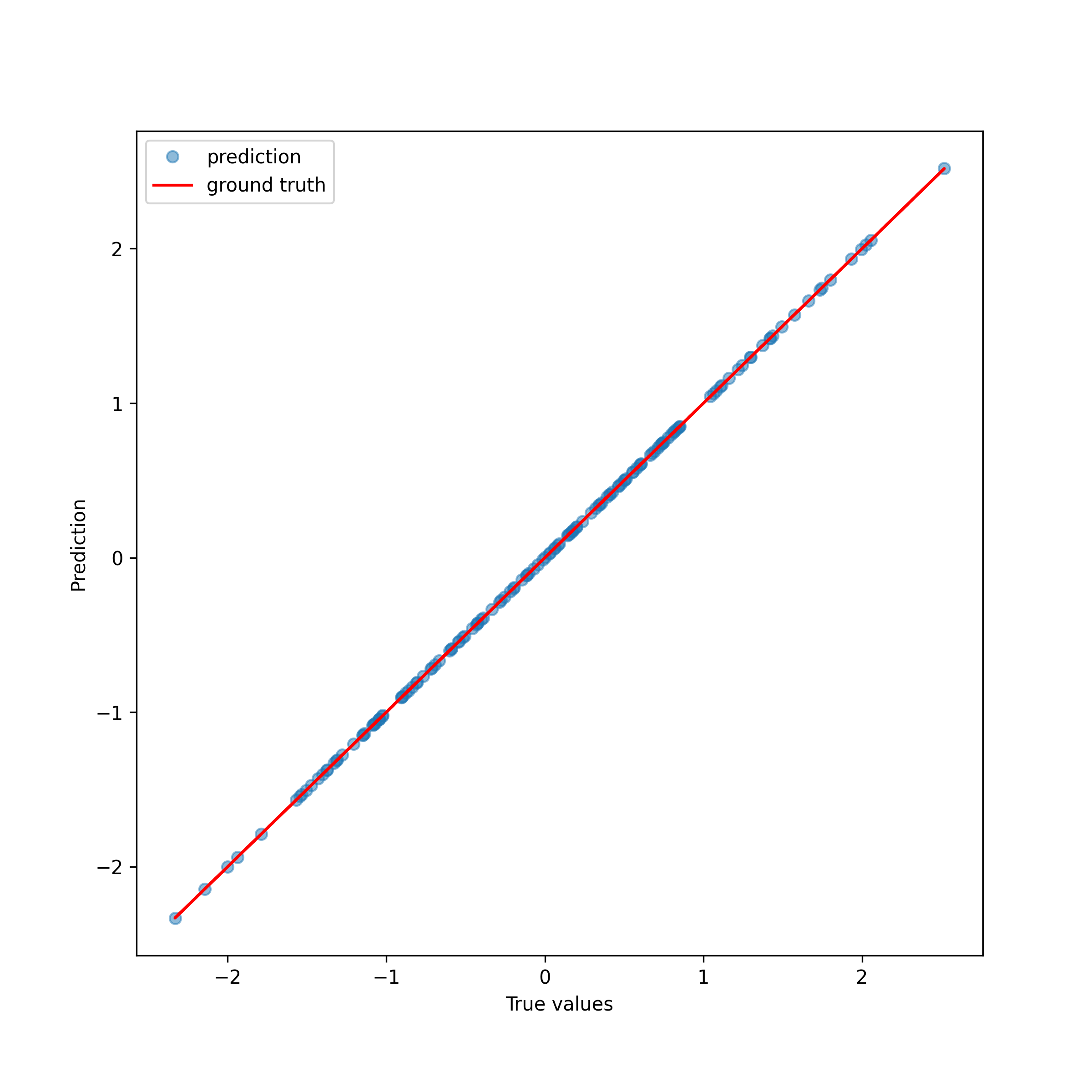}}
    % neural net ------------------------------------------------------------------
    \subfloat[NN, $\hat{\epsilon}_{\text{train}} = 0.0054$ \label{fig:performance-NN-crash}]{
        \includegraphics[width=.3\linewidth]{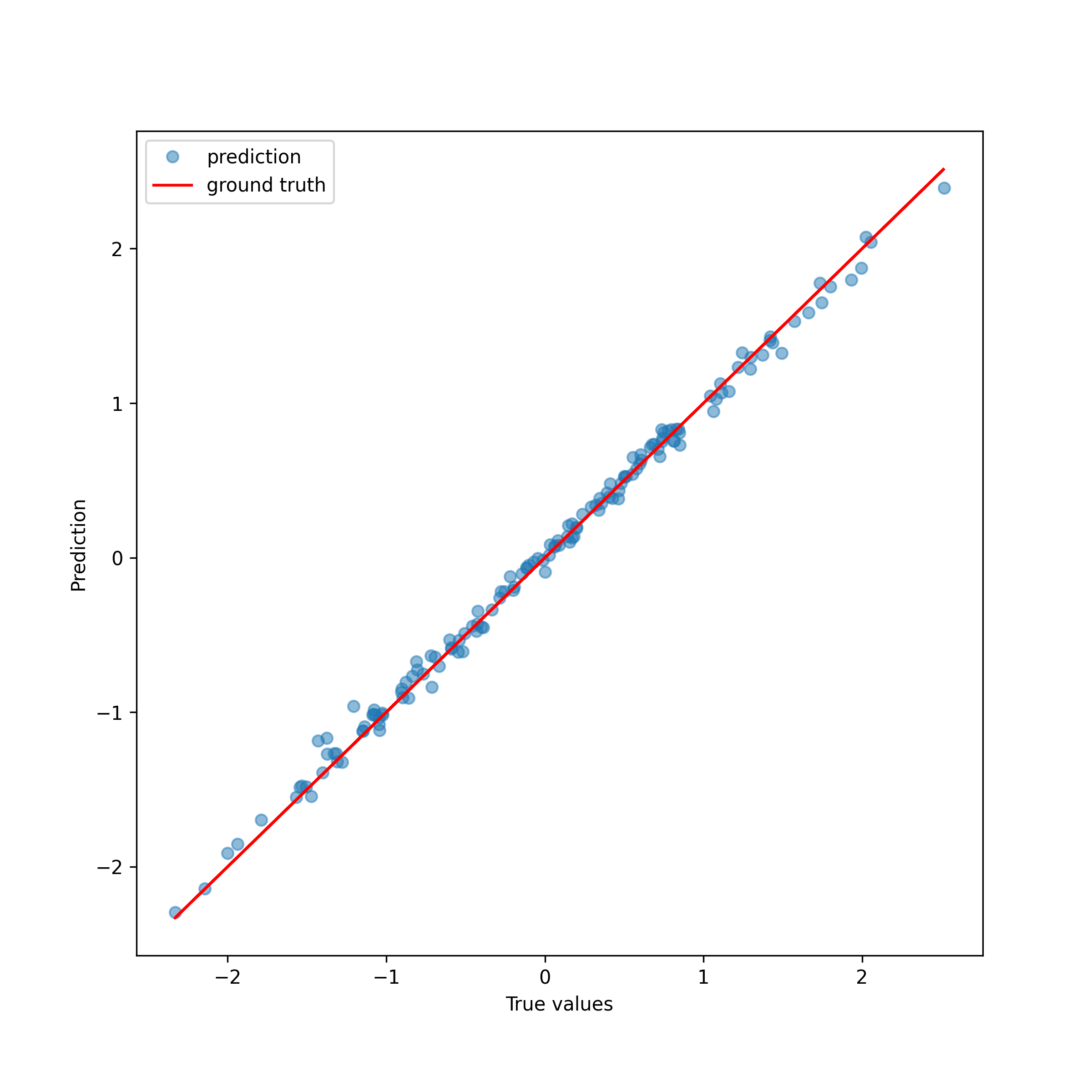}}
    % ours ------------------------------------------------------------------------
    \subfloat[LOS-RFE, $\hat{\epsilon}_{\text{train}}=0.0036$ ]{
        \includegraphics[width=.3\linewidth]{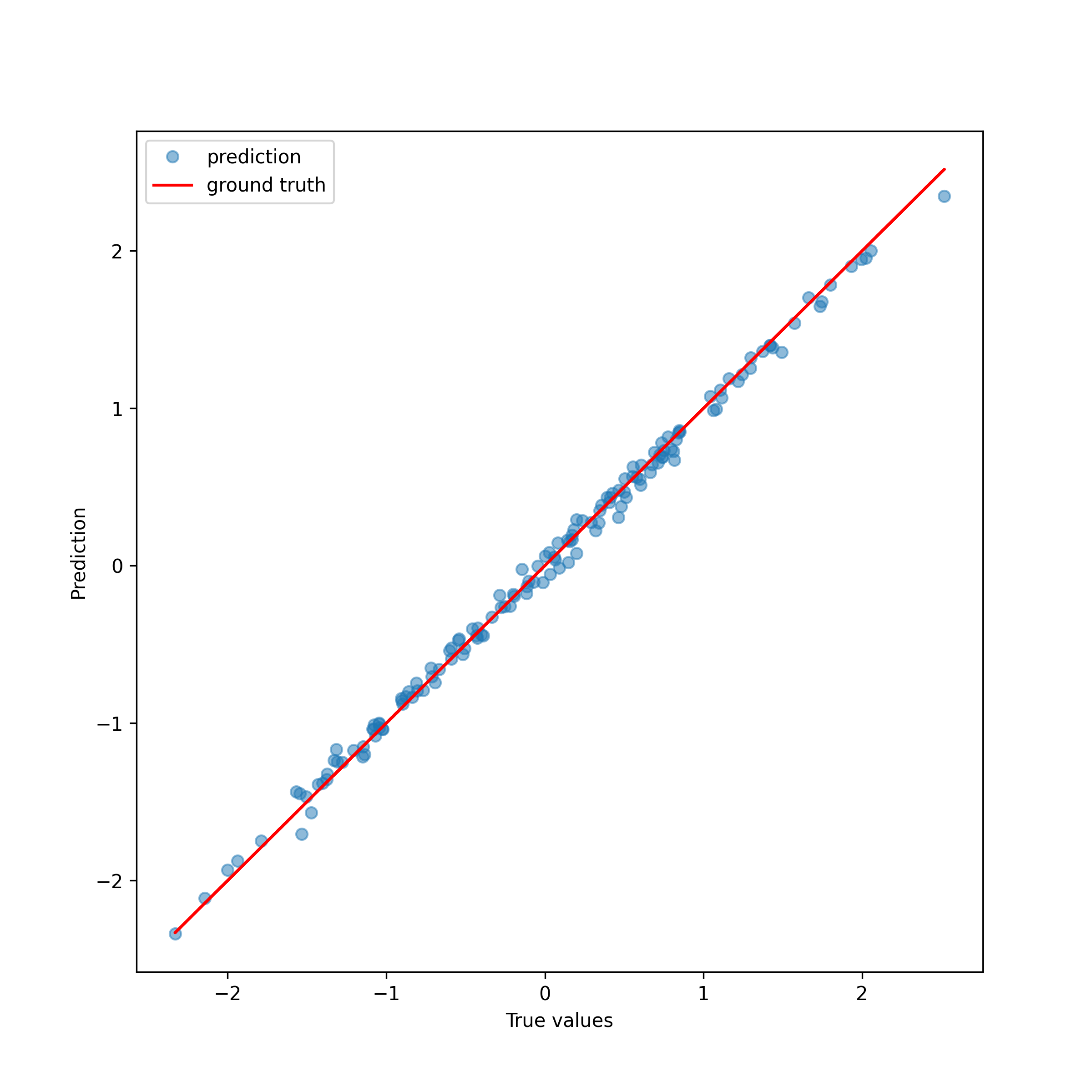}} 
    
    \vspace{0.2cm}
    
    % pce ------------------------------------------------------------------------
    \subfloat[PCE-based, $\hat{\epsilon}_{\text{test}}=0.0468$ \label{fig:performance-PCE-based-crash}]{
        \includegraphics[width=0.3\linewidth]{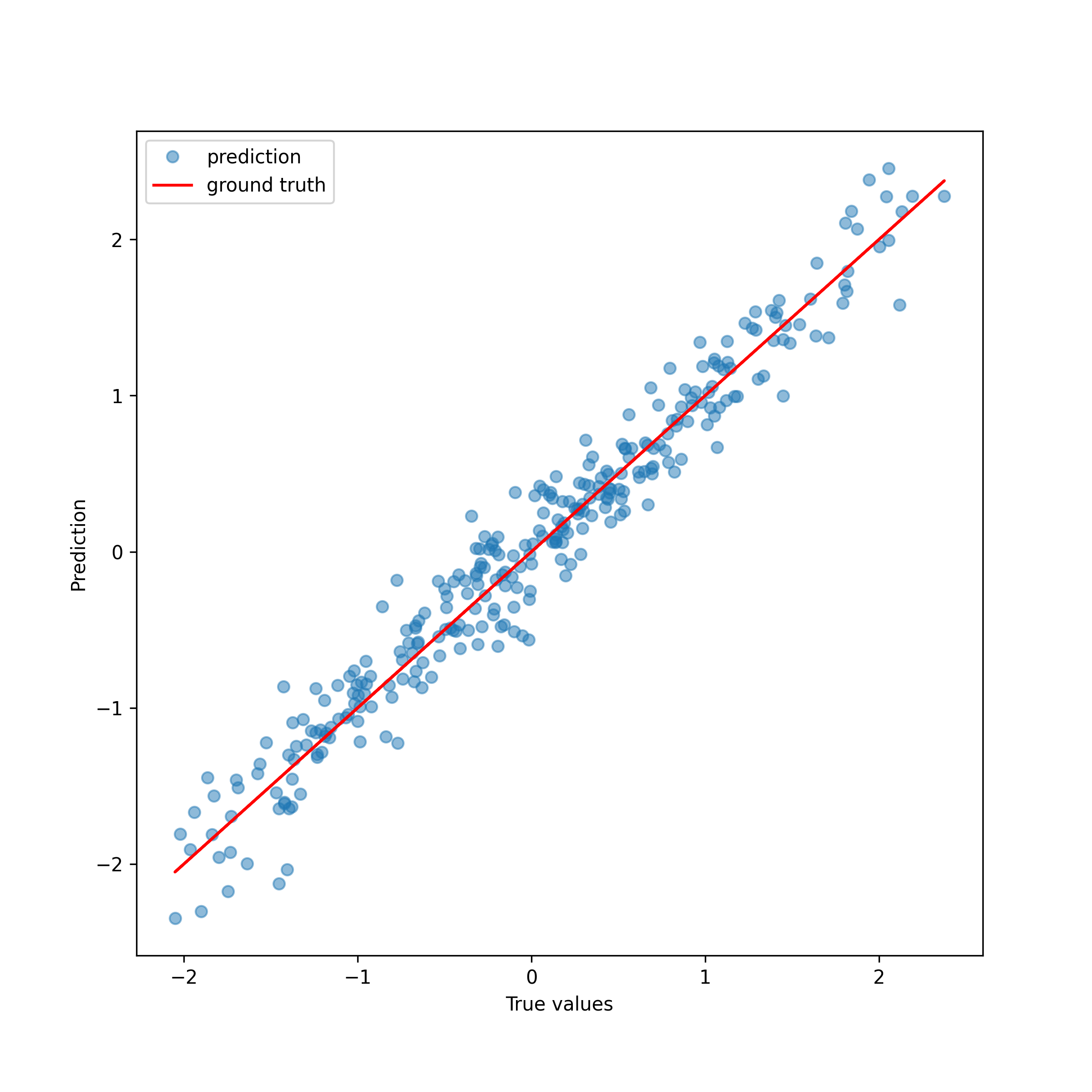}}
    % ----------------------------------------------------------------------------
    % nn -------------------------------------------------------------------------
    \subfloat[NN, $\hat{\epsilon}_{\text{test}}=0.0539$ \label{fig:performance-NN-crash-test}]{
            \includegraphics[width=.3\linewidth]{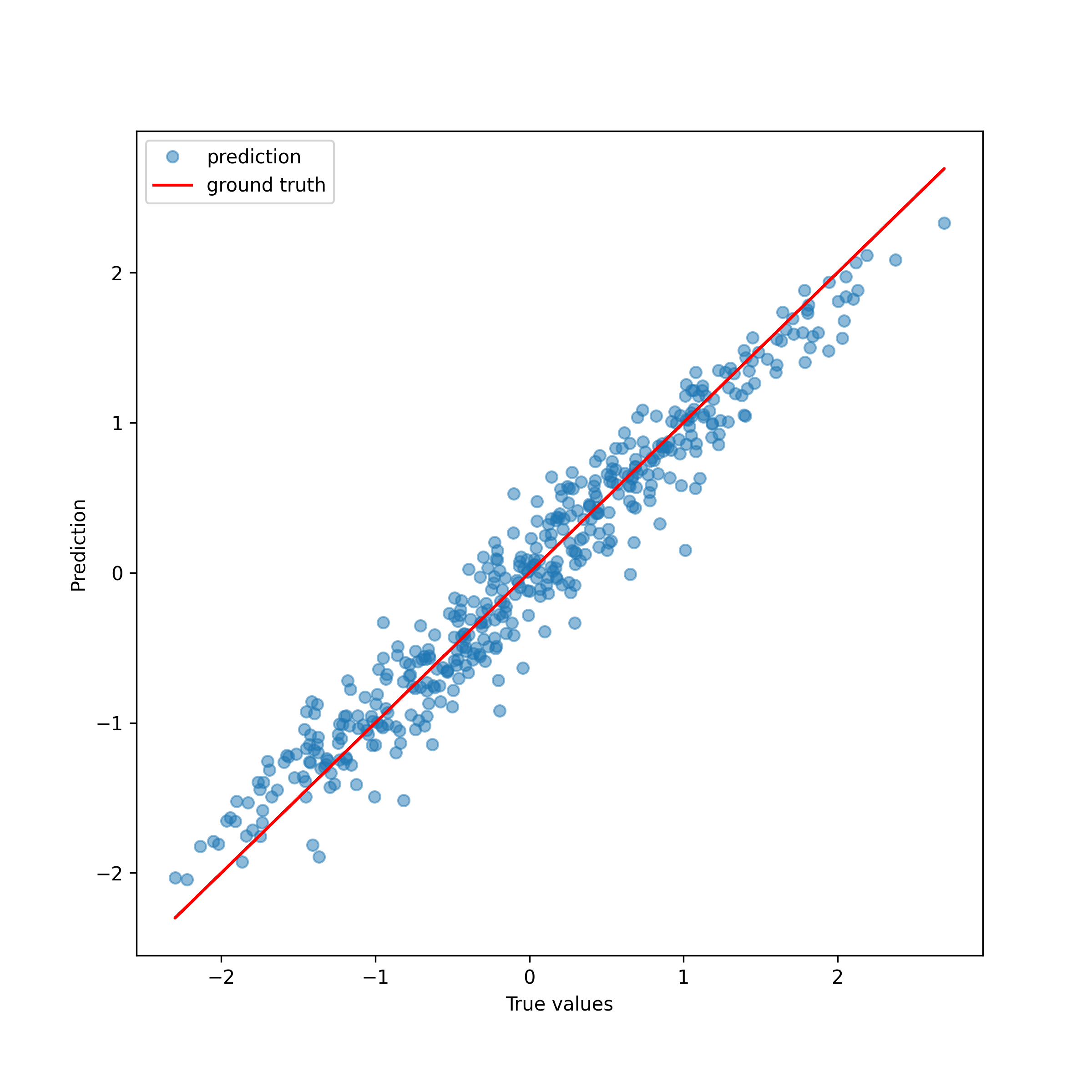}}
    % ----------------------------------------------------------------------------
    % ours -----------------------------------------------------------------------
    \subfloat[LOS-RFE, $\hat{\epsilon}_{\text{test}}=0.0186$]{
            \includegraphics[width=.3\linewidth]{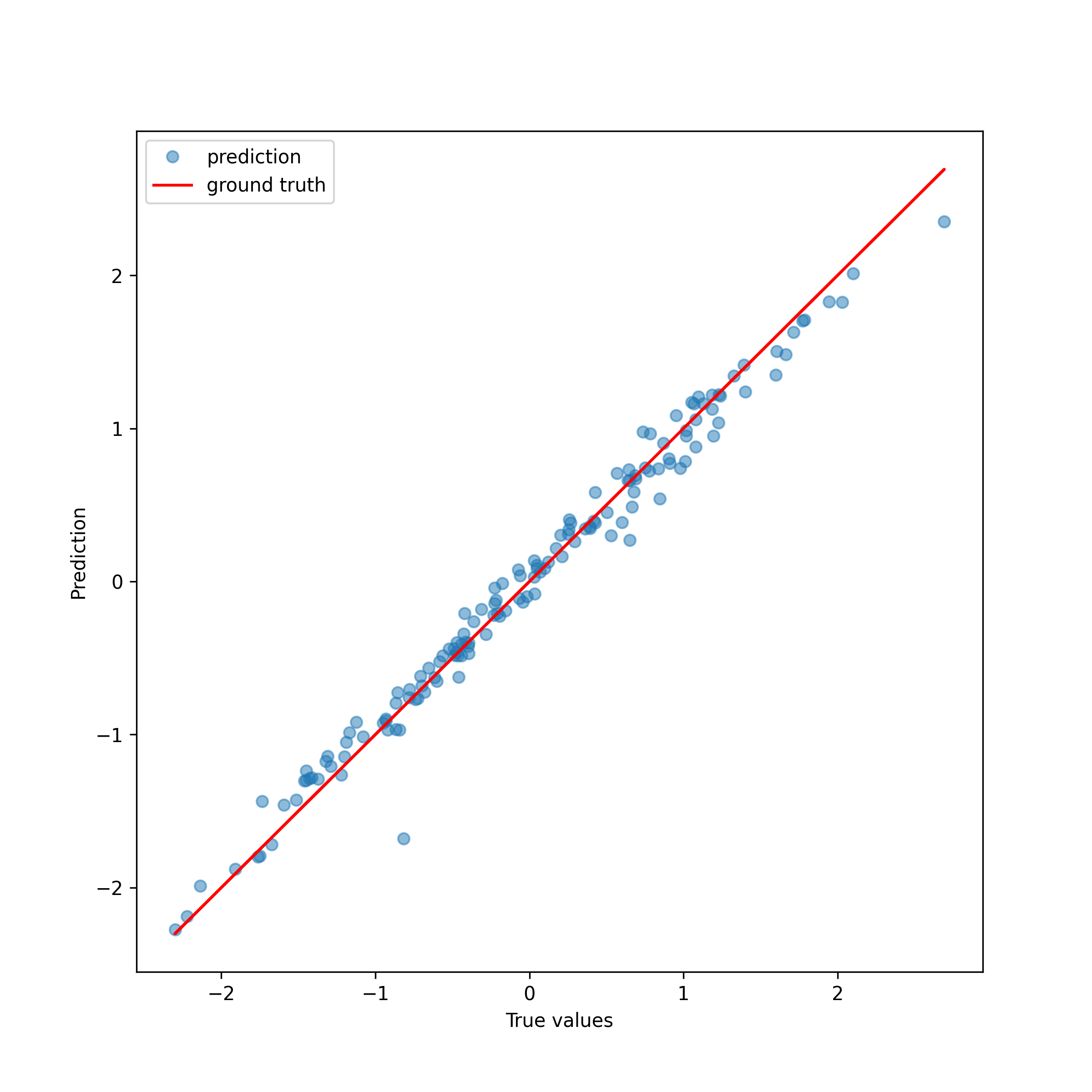}
            \label{fig:performance-srfe-crash-test}}
    % ----------------------------------------------------------------------------
    \caption[Performance of dimension reduction and PCE on crash test data]{
     The predictions depicted against the true values for the crash dataset for the PCE-based model (Figure~\ref{fig:performance-PCE-based-crash}), LOS-RFE (Figure~\ref{fig:performance-srfe-crash-test}) and the neural network (Figure~\ref{fig:performance-NN-crash-test}). 
        % For the latter one, the training and validation history shows a converged training without overfitting. 
        The loss of the best PCE-based model on the test data, with a latent dimension of $k=22$, is $\hat{\epsilon} = 0.0468$ on the test data. For the proposed LOS-RFE model, the latent dimension was chosen to be $k=14$ with a loss of $\hat{\epsilon} = 0.0186$. 
        LOS-RFE gives rise to the best  generalization error among the three methods.}
\end{figure}

The PCE-based model was examined first, showing a preferred latent dimension larger than $10$, as Figure~\ref{fig:performance-hist-pce-and-ours-on-crash-data} shows. While the optimization seems to have plateaued for the $k \in \{12, 14, 18, 20\}$, the optimization for $k=16$ might have reached the maximum number of iterations rather than a local minimum. As the results for the dimensions before and after $k=16$ seem to be far lower, this can be assumed to be an outlier rather than a single dimension yielding a bad fit. The best performance was obtained for $k=22$.
% while the results for $k=14$, $k=18$ and $k=20$ were only $0.001$ higher. 
The performance of the best PCE-configuration on the test data is shown in Figure~\ref{fig:performance-PCE-based-crash}, achieving the generalization  error $\hat{\epsilon} = 0.0.0468$ on the test set. Note, by looking at the training performance, the overfitting of the training data by the PCE-model can be identified. This is most likely the reason for the poor performance on the test data, compared to the good training performance. 
% ----------------------------------------------------------------------------
% performance of PCE and ours approach in the different latent dimensions

% ----------------------------------------------------------------------------

% ----------------------------------------------------------------------------
% best PCE-based, nn, and ours model on testing data
% ----------------------------------------------------------------------------

Second, the neural network with the same configuration as before is fit on the crash dataset. In contrast to the first experiment on Sobol data, the training dataset is far smaller leading to difficulties generalizing to unseen data. As Figure~\ref{fig:performance-NN-crash} shows, the training of the network has converged without overfitting. However, while the model seems to be robust, the evaluation on the hold-out test dataset reveals difficulties to predict values especially at the lower end of the range, see Figure~\ref{fig:performance-NN-crash-test}. This time, the performance of the neural network decreased with a minimum error of $\hat{\epsilon} = 0.0539$ on the test data.

% {\mr Third, the LOS-RFE  description: $R=10000 , \lambda=10^{-4}, q=2, $ }

Finally, the proposed method LOS-RFE  with $R=10000$ random features and sparsity level $q=2$ is determined using the training and validation sets of the crash test data. Similar to the PCE-based method, a favorable latent dimension resides in the last quarter in the range, taken into consideration. The dimension yielding the lowest error is $k=14$. An overview of the model's performance with respect to  the different latent dimensions is given in Figure~\ref{fig:performance-hist-pce-and-ours-on-crash-data}. The model resulting from the best configuration achieved an empirical generalization error of $\hat{\epsilon} = 0.0186$ on test data, which is again lower then the errors for the neural networks as well as the PCE-based method.

We would like to note that the proposed LOS-RFE can have a very high complexity by choosing a large number of random features. This can, similar to too large neural networks, lead to overfitting the data. To demolish the overfitting effect, special care has to be taken how the hyper-parameters $R$, the number of random features, sparsity $q$ and $\lambda$, the scale of the regularizer, are chosen.

It shall also be noted, that the LOS-RFE method is more resource-demanding as it requires more memory when creating the large matrices, holding the random features. Also, due to the large coefficient vector, the time to find a solution to the LASSO problem usually takes more time than solving the minimization problem for the PCE variant. However, the resource consumption and compute time when fitting a PCE depends directly on the maximum order of the monomials used for the expansion. 

However, a great amount of computation is still spent on finding a suitable lower dimensional representation of the data, which is shared by both methods, PCE-based and the proposed approach. Incorporating methods for estimating the latent dimension would decrease the runtime drastically, as the optimizer for the reduction method would only have to be run once.

Finally, the overall results on the test data for experiments carried in this section are summarized in Table~\ref{tab:table-of-results}.

\begin{table}[ht]
    \centering
    \begin{tabular}{r|c|c|c|c}
                            & \multicolumn{3}{c}{Test error $\hat{\epsilon}$} \\ \hline
        \textbf{Model}      & \textbf{Sobol data} & \textbf{inc. Sobol data}   & \textbf{Crash test data} \\ \hline \hline
        PCE-based           & $0.0171$      & .         & $0.0468$       \\
        Neural Net.         & $0.0151$      & $0.0034$  & $0.0539$       \\
       LOS-SRF              & $0.0062$      & $0.0052$  & $0.0186$               
    \end{tabular}
    \caption[Final results obtained from the experiments]{Final results obtained from the experiments described above, where the fitted model was evaluated on test data, which was reduced to the most suitable latent dimension using the according $\theta$.}
    \label{tab:table-of-results}
\end{table}

% -------------------------------------------------------------------\newpage-----
% -------------------------------------------------------------------\newpage-----

\newpage
\subsection{Summary}

The examples above have shown the high accuracy of the proposed method of a self-supervised combination of dimensionality reduction and LOS-RFE. Also, the advantage of a self-supervised loop is observable by the streamlined process with no further interaction by the applicant than providing the input data. Compared to state-of-the-art techniques for surrogate modelling, such as  PCE-based methods and neural networks, the proposed LOS-RFE method has shown clear advantages in the scarce data setting. 

While the networks learn the parameters for all its nodes, the LOS-RFE approach learns to use only the most informative features. This is the reason why surrogate modelling based on random features has an advantage in data-scarce settings, compared to neural networks, which are data hungry as it was also mentioned in \cite{hashemi2021generalization}.

% -----------------------------------------------------------------------------
\section{Conclusion}\label{sec:conclusion}

In the discipline of UQ, the goal is to measure the uncertainty within a simulation or computational model. As most such computational models are too complex to be run numerous times to construct a proper distribution of the output, conditioned on the input, surrogate models are used as a fast-to-evaluate replacement. This also implies that only few evaluations from the computational model are available
to design such surrogates. For the surrogate models, it is essential to approximate the computational model with high accuracy.

In this work, a method of 
construction of high-fidelity surrogates in data-scarce settings has been proposed. Namely, it has been shown that low-order sparse random feature expansion (LOS-RFE) can be used to surrogate models of higher accuracy compared to state-of-the-art surrogate modelling
techniques. Further, a self-supervised process including dimensionality reduction has been established with the aforementioned model by using the model's performance as feedback for the reduction method. This allows for optimizing over the parameters of the reduction as well as the surrogate's parameters with respect to data representation. In addition, such  representation of the data in lower dimensions was learned to be
best-suited for the LOS-RFE surrogate model, possibly increasing the surrogate model's performance.

As an object for future work, we see further analysis of different metaheuristics used to optimize the reductions parameter's as well as the reduction method itself. Regarding the reduction method, the here used Kernel PCA benefits  from the small dataset and is thus very performant. In the case of larger datasets, KPCA will turn to be a bottleneck due to increased memory requirements. Finally, it could also be interesting to see how the method performs on larger datasets as the sparsity-of-effects principle is often stated for such as well.

Last but not least, it has to be mentioned that the method of LOS-RFE requires a certain amount of memory. This is due to the large matrices
holding several features per observation. Thus, given a dataset of fixed size $N$, the memory requirement grows with the order of the random features $\mathcal{O}(R\cdot N)$.  Therefore, when using LOS-RFE one should mainly focus on scarce data settings, which was also the main motivation for our work. While most machines allow running the application using multithreading or even multiprocessing, the available memory is the bottleneck. This is in contrast to other surrogate models such as PCE, which require more resources on the CPU side. However, the datasets in the described setting of UQ are relatively small and thus memory size of today's high-end personal computers should suffice in most of the applications. 

Concluding, we have shown in this work how a method such as LOS-RFE, inspired by compressive sensing, can be successfully used for surrogate modelling. When further coupled with dimensionality reduction in a self-supervised manner, the presented algorithm yields very good performance, with a minimal set of hyper-parameters to be determined.

% -----------------------------------------------------------------------------
% Acknowledgments
% \input{acknowledgments}

\paragraph*{ \large Acknowledgments.} FK and AV acknowledge support by the German Science Foundation (DFG) in the context of the Emmy Noether junior research group KR 4512/1-1 and the collaborative research center TR-109 as well as by the Munich Data Science Institute and Munich Center for Machine Learning. MH and JJ acknowledge support by the BMW Group in the context of the ProMotion program. 
The authors would like to thank Marco Rauscher for  valuable discussions related to the numerical experiments.

% -----------------------------------------------------------------------------
% \appendix
% \textcolor{red}{do we need an appendix? \textemdash probably we don't}
% \include{chapters/appendices_all}

\end{document}